\documentclass{article}

\PassOptionsToPackage{numbers}{natbib}
\usepackage{neurips_2025}

\usepackage[utf8]{inputenc} 
\usepackage[T1]{fontenc}    
\usepackage{hyperref}       
\usepackage{url}            
\usepackage{booktabs}       
\usepackage{amsfonts}       
\usepackage{nicefrac}     
\usepackage{microtype}   
\usepackage{xcolor}       
\usepackage{graphicx}      
\usepackage{amsmath}  
\usepackage{amssymb}   
\usepackage{bm}        
\usepackage{tablefootnote}
\usepackage{booktabs}
\usepackage{colortbl}
\usepackage[ruled]{algorithm2e}
\usepackage{multirow}
\usepackage{enumitem}  
\usepackage{siunitx}      

\title{%
  \raisebox{-0.4\height}{\includegraphics[height=1cm]{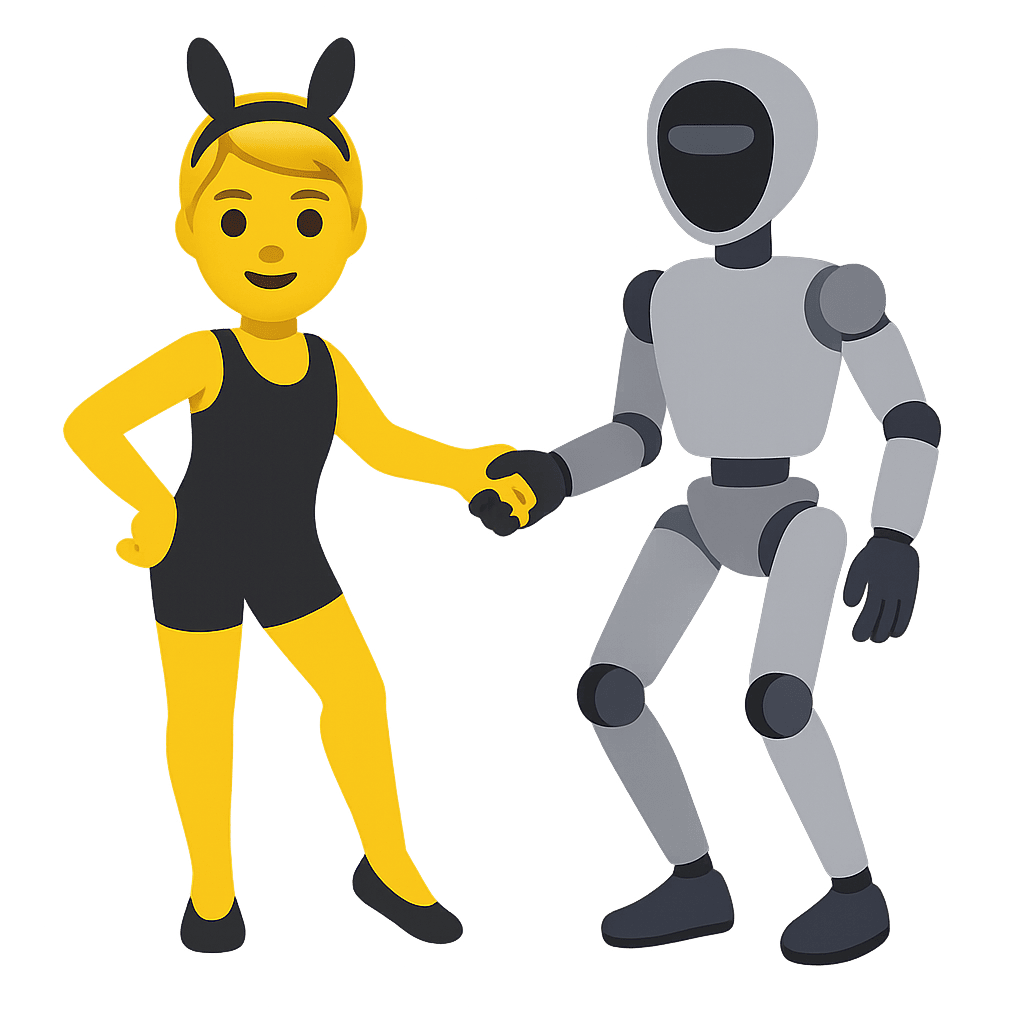}}
  \hspace{0.1em} 
  DanceTogether! Identity‑Preserving Multi‑Person Interactive Video Generation
}

\begin{document}

\maketitle               

\begin{center}
    \vspace{-1cm}
    Project Page: \url{https://DanceTog.github.io/} \\
    \vspace{0.5cm}
\end{center}

\begin{figure}[h]
\vspace*{-2em}
\centering
\includegraphics[width=\textwidth]{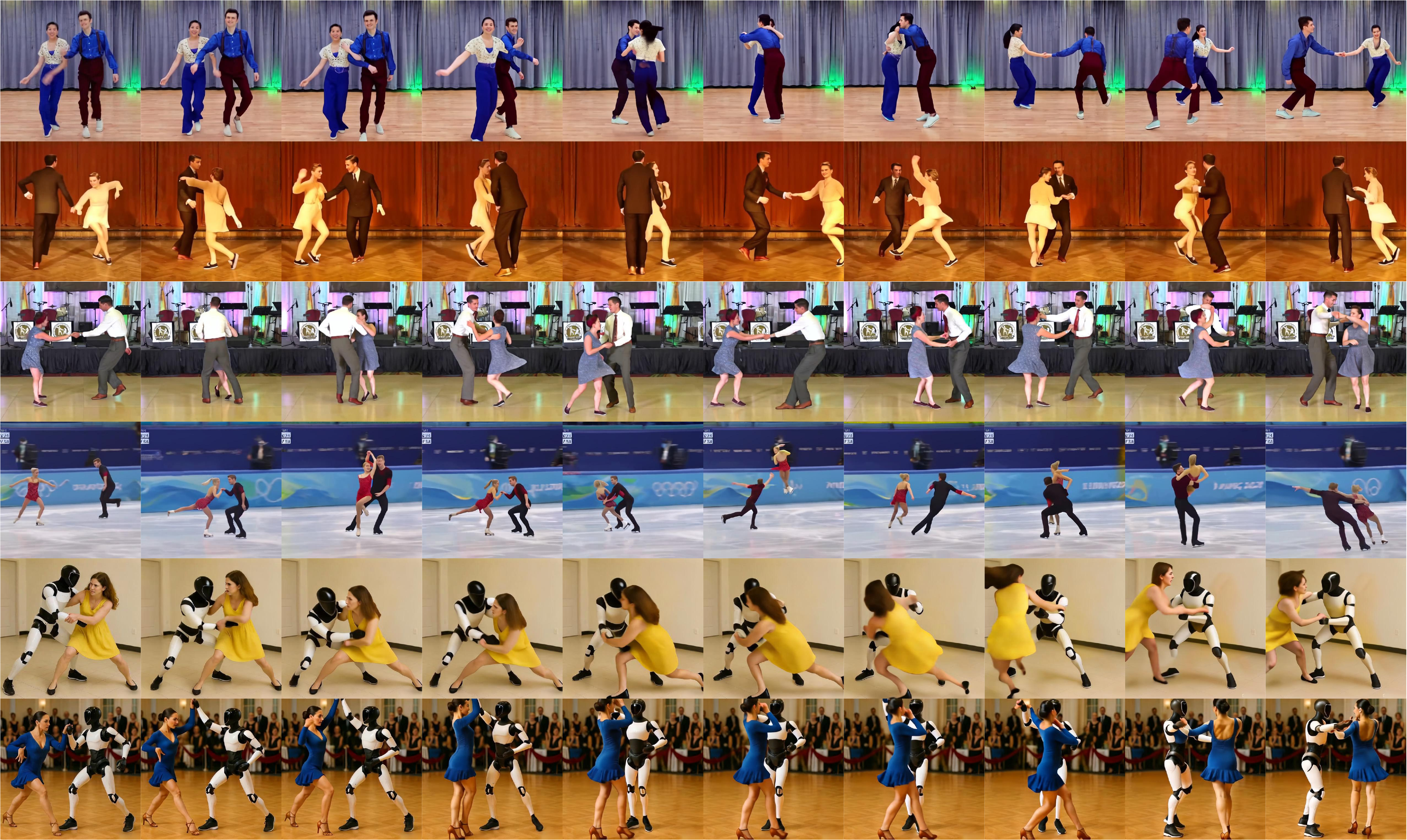}
\caption{
\emph{DanceTogether} generates complex two-person interaction videos with interactive details and consistent identity preservation from a single reference image (see the left-most of each row), using independent multi-person pose and mask sequences as control signals. 
}
\label{fig:teaser}
\end{figure}

\renewcommand{\thefootnote}{} 
\footnotetext{
    \textsuperscript{\textdagger} Corresponding Author.
}
\renewcommand{\thefootnote}{\arabic{footnote}} 

\begin{abstract}
Controllable video generation (CVG) has advanced rapidly, yet current systems falter when more than one actor must move, interact, and exchange positions under noisy control signals. We address this gap with \emph{DanceTogether}, the first end-to-end diffusion framework that turns a single reference image plus independent pose--mask streams into long, photorealistic videos while \emph{strictly preserving every identity}. A novel \textit{MaskPoseAdapter} binds ``who'' and ``how'' at every denoising step by fusing robust tracking masks with semantically rich---but noisy---pose heat-maps, eliminating the identity drift and appearance bleeding that plague frame-wise pipelines. To train and evaluate at scale, we introduce (i) \texttt{PairFS-4K}, \(26\,\text{h}\) of dual-skater footage with $7{,}000+$ distinct IDs, (ii) \texttt{HumanRob-300}, a one-hour humanoid--robot interaction set for rapid cross-domain transfer, and (iii) \texttt{TogetherVideoBench}, a three-track benchmark centred on the \texttt{DanceTogEval-100} test suite covering dance, boxing, wrestling, yoga, and figure skating. 
On \texttt{TogetherVideoBench}, \emph{DanceTogether} outperforms the prior arts by significant margin.
Moreover, we show that a one-hour fine-tune yields convincing human--robot videos, underscoring broad generalization to embodied-AI and HRI tasks. 
Extensive ablations confirm that persistent identity--action binding is critical to these gains. 
Together, our model, datasets, and benchmark lift CVG from single-subject choreography to \textbf{compositionally controllable, multi-actor interaction}, opening new avenues for digital production, simulation, and embodied intelligence.
Our video demos and code are available at \url{https://DanceTog.github.io/}.
\end{abstract}

\section{Introduction}
\emph{Controllable video generation} (CVG) ~\cite{xue2024human,wang2023motionctrl,ma2024follow,kuang2024collaborative,peng2024controlnext} seeks to translate explicit control signals—\emph{e.g. }per-frame human poses, body masks, or trajectory commands—into photorealistic human-motion videos. 
Compared to AI generation tasks that use single conditioning (reference images or text)\cite{podellsdxl, blattmann2023stable, sora2024, khachatryan2023text2video}, some controllable generation tasks typically combine multi-modal conditions as input~\cite{zhang2023adding, peng2024controlnext, shi2024motion, hu2024animate, yang2024idea2img, chen2023control3d, chen2024idea, ma2024follow}. Such tasks using multi-modal control signals have broad and important applications in film production~\cite{bugliarello2025you, song2024directorllm, hu2022make}, digital human interaction~\cite{tian2025emo2,xu2024xagen,chen2024ultraman,yan2024dialoguenerf,sun2024drive,lee2002interactive,wang2025instructavatar}, and embodied AI~\cite{bharadhwaj2024gen2act, wang2024language, cai2018deep,zhao2025taste, krumme2025world, hao2018controllable,qin2024worldsimbench,xue2024human,duan2022survey,yang2024holodeck,albaba2025nil}.
In particular, we investigate the task of CVG with multi-person interactions, which is highly challenging as it simultaneously requires
(i) \textbf{preserve the identities of multiple actors} over hundreds of frames,
(ii) \textbf{maintain the spatio-temporal coherence of complex interactions} such as hand-holding, lifts, position exchanges, and synchronous choreography, and
(iii) \textbf{faithfully obey noisy control signals} in the presence of occlusion, motion blur, and rapid viewpoint changes.

Most existing systems adopt a frame-wise synthesis followed by temporal smoothing paradigm: each image is generated independently from pose or text conditions and then stitched into a video via interpolation, optical-flow warping, or temporal convolutions \cite{di2023magicdance,ma2024follow,zhu2024generative,chan2019everybody,yang2018pose,luo2023notice}. Nearly all of these models are trained solely on single-person dance datasets \cite{zhu2024champ,xu2023magicanimate,hu2024animate,ma2024follow,wang2025unianimate,mimicmotion2024,Wang_2024_CVPR,karras2023dreampose,wang2024vividpose,li2025dispose,peng2024controlnext}. A handful of works incorporate multi-person footage \cite{wang2024humanvid,zhang2024follow,xue2024follow2}, but they exhibit pronounced \emph{identity drift} and appearance bleeding when the actors exchange positions. In general, state-of-the-art methods struggle with identity inconsistency, cross-subject contamination, and missing interaction details—issues that rapidly worsen once more than one performer is involved. 

We present \emph{DanceTogether}, the first end-to-end diffusion framework expressly tailored for controllable multi-person interaction video generation. Our guiding hypothesis is that robust multi-actor synthesis requires an \emph{explicit, persistent binding between identity and motion} throughout the diffusion process. To this end, we deliberately disentangle identity from action and then re-couple them: instead of relying solely on fragile pose estimates, we fuse stable tracking masks with semantically rich pose cues. This fusion is realised by a novel conditional adapter, MaskPoseAdapter, which combines the \emph{reliable, easy-to-obtain body masks} with the \emph{informative yet noisy poses} into a bimodal control signal. By integrating each subject’s mask and pose into a unified representation, the adapter enforces precise identity-to-action alignment at every generative step.

Our framework operationalizes the identity–action binding principle through three tightly coupled modules. (i) MultiFace Encoder distills a compact set of identity tokens from a single image and injects them into every cross-attention layer, ensuring subject appearance is held constant throughout the sequence. (ii) MaskPoseAdapter fuses robust per-person tracking masks with semantically rich—but noisy—pose maps to deliver a bimodal conditional signal that aligns “who” and “how” at every diffusion step, thereby safeguarding both identity integrity and motion fidelity. (iii) Video Diffusion Backbone leverages these aligned signals to synthesize high-resolution clips whose multi-actor motions remain coherent, physically plausible, and free of inter-subject drift. 

Extensive evaluation on the new TogetherVideoBench—built around our 100-clip DanceTogEval-100 set—shows that DanceTogether decisively advances controllable multi-person video generation. Across the three core tracks (Identity-Consistency, Interaction-Coherence, Video Quality) it raises the bar over the strongest prior (StableAnimator~\cite{tu2024stableanimator} +swing dance data~\cite{swingdance} finetune) by +12.6 HOTA, +7.1 IDF1, +5.9 MOTA, trims MPJPE$_{2D}$ by 69 \% (1555 → 492 px), and boosts OKS/PoseSSIM to 0.83/0.93. 
Visual fidelity also improved accordingly: human mask region FVD/FID decreased from 29.0/66.7 to 17.1/48.0, without sacrificing CLIP alignment effect.
Fine-tuning on our proposed one-hour HumanRob-300 dataset can generate convincing human-robot interaction videos, which highlights the framework's broad generalization capability and prospects in embodied AI research.

To summarize, our main contributions include: 

\begin{enumerate}[leftmargin=1.4em,itemsep=1pt]
\item \textbf{DanceTogether framework.}
We present the first end-to-end diffusion framework for controllable multi-person interaction video generation. Our novel \emph{MaskPoseAdapter} fuses stable tracking masks with pose cues to enforce identity-action binding throughout the generation process.

\item \textbf{Data curation pipeline and datasets.}
We develop a monocular-RGB pipeline for extracting tracking-aware human poses and masks. Using this, we curate \texttt{PairFS-4K} (26h dual-person figure skating) and \texttt{HumanRob-300} (1h robot interaction) datasets.

\item \textbf{TogetherVideoBench benchmark.}
We introduce a comprehensive evaluation benchmark with three tracks (\emph{Identity-Consistency}, \emph{Interaction-Coherence}, \emph{Video Quality}) and \texttt{DanceTogEval-100} containing 100 dual-actor clips across diverse activities.

\item \textbf{Superior performance and generalization.}
Our method achieves significant improvements: +12.6 HOTA, +7.1 IDF1, +5.9 MOTA over the strongest baseline, 69\% reduction in pose error, and enhanced visual fidelity (FVD: 29.0→17.1). Cross-domain fine-tuning demonstrates strong generalization to human-robot scenarios.
\end{enumerate}
\section{Related Work}
\subsection{Diffusion Models for Video Generation}

In recent years, diffusion models have achieved great achievements in the field of video generation~\cite{poole2022dreamfusion,li2023finedance,kong2024hunyuanvideo,li2024lodge,li2024lodge,wang2024cove,wang2024taming,chen2024follow,lin2024open,han2024reparo}. In the technical solution of video generation model, early work mainly used 3D-Unet to achieve consistent fusion of time and space~\cite{singer2022make,ho2022video}. On this basis, ~\cite{blattmann2023align} introduces the time dimension into the latent spatial diffusion model to convert the image generator into a video generator; further, ~\cite{ho2022video} uses the basic video generation model and a series of interleaved spatial and temporal video super-resolution models to generate high-definition videos; ~\cite{xu2024easyanimate} is based on end-to-end video generation and editing of the diffusion model, and uses spatiotemporal consistency modeling and multimodal condition control to support video generation under multimodal conditions such as text, images, and video. ~\cite{blattmann2023stable} Based on the potential diffusion model transformation of 2D image synthesis training, a good time insertion strategy for managing video data is proposed. 
Although large-scale commercial pre-trained models such as ~\cite{kong2024hunyuanvideo,kling2024,sora2024} have good time consistency and high resolution, they still cannot meet the video generation task using fine human motion control signal input. 

\subsection{Controllable Human Video Generation}
The integration of diffusion models~\cite{latentdiffusion,blattmann2023stable,ma2024follow,xue2024follow2,zhang2024follow,wang2024vividpose,xu2024magicanimate,hu2024animate,li2025magicmotion,li2025dispose,karras2023dreampose,feng2023dreamoving,Wang_2024_CVPR,tu2024stableanimator,zhu2024champ,peng2024controlnext} has greatly advanced controllable human video generation, with most methods building on pre-trained Stable Diffusion and incorporating action or pose guidance for continuous video synthesis. Pose conditions are commonly represented by keypoints or skeletons, as in ControlNet~\cite{zhang2023adding} and ReferenceNet~\cite{hu2024animate}, and are used as conditional inputs during denoising. For instance, Disco~\cite{Wang_2024_CVPR} separates background and pose control via dedicated modules, a strategy extended by later works~\cite{hu2024animate,xu2024magicanimate} to improve video continuity. Other approaches~\cite{wang2024vividpose,lin2023rich,mimicmotion2024} introduce geometric priors, using rendered images from 3D models (e.g., depth, normal, semantic maps) as pose conditions, while methods like~\cite{zhu2024champ,li2025dispose} employ SMPL models or 2D keypoints, but are mostly limited to single-person or simple multi-person scenarios.
Despite these advances, most methods focus on single-person generation and struggle with complex multi-person interactions and identity consistency. To address identity preservation, recent works~\cite{wang2024humanvid,wang2025unianimate,mimicmotion2024,luo2025dreamactor,wang2025multi} explore pose-guided identity maintenance, such as using identity encodings or masks~\cite{yoon2021pose}, but these are often limited to short or simple videos. Tevet et al.~\cite{tevet2022human} generate high-quality action sequences but lack robust identity modeling for long, complex videos. Some video-oriented methods~\cite{wang2024vividpose,mimicmotion2024,wang2025multi} use local masks or attention to reduce identity confusion, but still lack explicit identity-action binding, leading to drift in long sequences.

\section{Method}
\subsection{Overview: DanceTogether Pipeline}

\begin{figure}[t!]
    \centering
    \includegraphics[width=\linewidth]{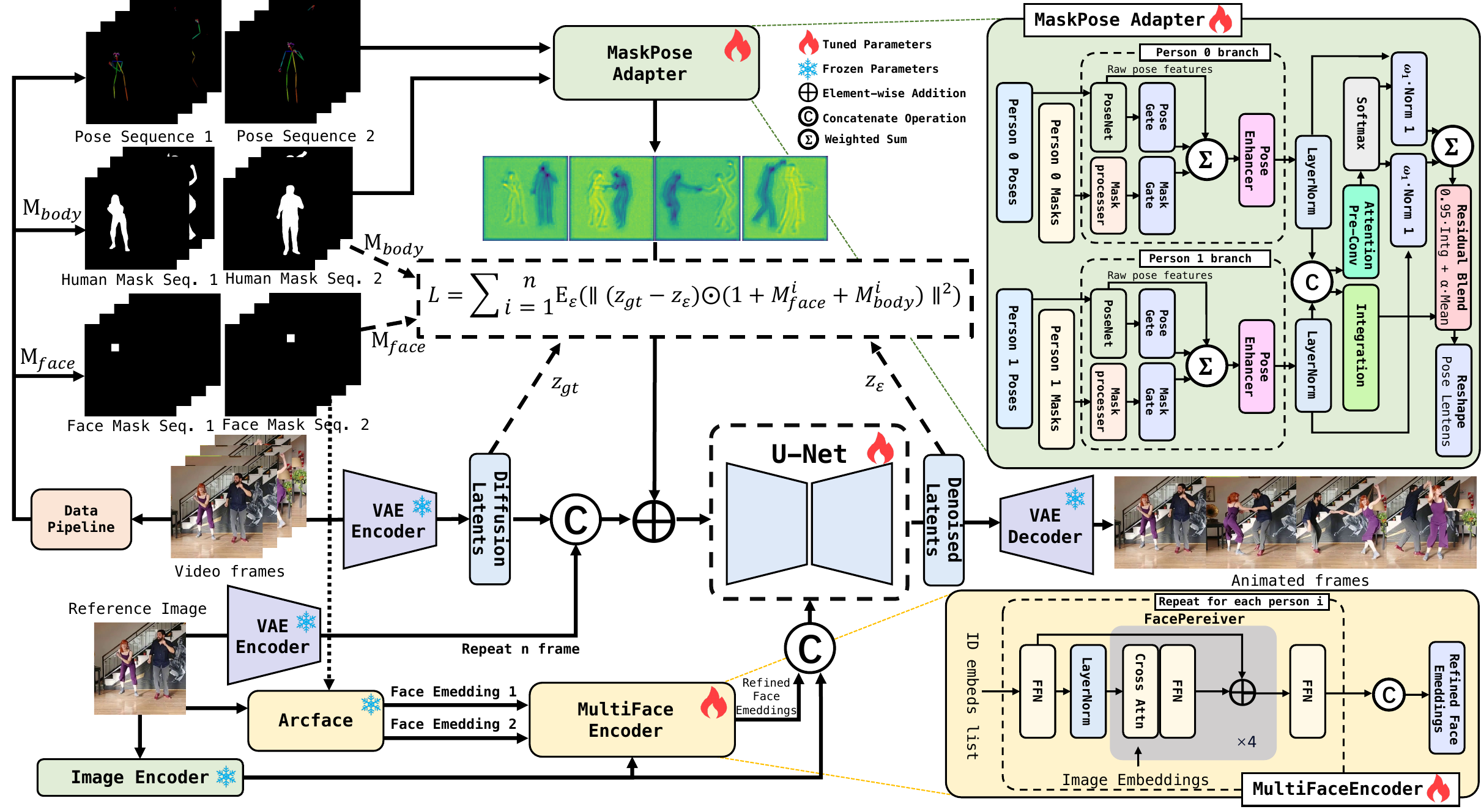}
    \caption{\emph{DanceTogether} pipeline overview: A single reference image and per-person pose/mask sequences enter the system; the MaskPoseAdapter fuses these control signals, the MultiFace Encoder injects identity tokens, and the video-diffusion backbone synthesizes an interaction video that preserves consistent identities for all actors.}
    \label{fig:pipeline_overview}
\end{figure}

Given a reference image $\mathbf{I}_{\text{ref}}$ and \textit{per-person} control signals $\{\mathbf{P}_i,\ \mathbf{M}_i\}_{i=1}^{N}$ (pose maps and tracking masks for $N$ individuals across $T$ frames), \emph{DanceTogether} synthesizes a video $\hat{\mathbf{V}}\in\mathbb{R}^{T\times3\times H\times W}$ while (i) preserving each identity, (ii) respecting the spatio-temporal interaction encoded in the poses, and (iii) remaining consistent with the poses and masks. 
The pipeline (Fig.~\ref{fig:pipeline_overview}) contains three key learnable modules including Video Diffusion Backbone (Sec.~\ref{sec:backbone}), MaskPoseAdapter (Sec.~\ref{sec:maskpose}) and MultiFace Encoder (Sec.~\ref{sec:fusionface}). 

\subsection{Video Diffusion Backbone}\label{sec:backbone}
\noindent\textbf{Starting point – \textit{StableAnimator}.}
Our backbone follows the \textbf{StableAnimator} architecture \cite{tu2024stableanimator}:  
a 16-frame latent UNet $f_{\theta}$ derived from Stable Video Diffusion (SVD).  
For every training clip we take as input
$\bigl(\mathbf{I}_{\mathrm{ref}},\ \mathbf{P}_{1:T},\ \mathbf{M}_{1:T}\bigr)$
where $\mathbf{I}_{\mathrm{ref}}\!\in\!\mathbb{R}^{3\times H\times W}$ is a reference image, and $\mathbf{P}_{t}$ / $\mathbf{M}_{t}$  are the pose map and tracking mask at frame $t$.

\noindent\textbf{Three conditioning streams.}
The UNet is conditioned by three streams, each of which begins with a \emph{frozen} pretrained encoder and is then refined by trainable adapters (see Fig.\,\ref{fig:pipeline_overview}):

\begin{itemize}[leftmargin=1.4em,itemsep=1pt]
  \item \textbf{Latent image stream.}  
        A frozen SVD VAE encoder maps both the reference image \(\mathbf{I}_{\mathrm{ref}}\) and each input video frame to their respective latent representations. 
        The reference latent \(\mathbf{z}_{\mathrm{ref}}\in\mathbb{R}^{C\times64\times64}\) is tiled along the temporal axis and concatenated with the per-frame latents \(\mathbf{z}_{gt}\).  
        This concatenated tensor is then fused with the \emph{trainable} MaskPoseAdapter’s condition latents via element-wise addition, producing the final latent input to the UNet.
  \item \textbf{CLIP image embeddings.}  
        A frozen ViT-H/14 encoder \(\phi_{\mathrm{CLIP}}\) produces 
        \(\mathbf{e}^{\mathrm{clip}}\!\in\!\mathbb{R}^{1024}\).  
        These embeddings serve as keys/values in every \emph{trainable} cross-attention block.
  \item \textbf{Refined face embeddings.}  
        A frozen ArcFace model \(\phi_{\mathrm{ID}}\) outputs 
        \(\mathbf{e}^{\mathrm{id}}\!\in\!\mathbb{R}^{512}\),  
        which is then refined by the \emph{trainable} MultiFaceEncoder \(g_{\psi}\):  
        \begin{equation}
          \mathbf{E}^{\mathrm{face}}
          = g_{\psi}\!\bigl(\mathbf{e}^{\mathrm{id}},\ \mathbf{e}^{\mathrm{clip}}\bigr)
          \in\mathbb{R}^{K\times d},
          \label{eq:face_tokens}
        \end{equation}
        implemented as four Perceiver-IO layers (\(K=4,\ d=768\)).  
        The resulting identity tokens modulate the same trainable cross-attention layers.
\end{itemize}

\noindent\textbf{Distribution‐aware ID Adapter.}  
To prevent a feature‐distribution shift when injecting identity tokens, StableAnimator inserts an ID Adapter before each temporal block.  Given input features \(\mathbf{h}\), we first apply spatial self‐attention and two cross‐attention steps, then align and fuse the face branch to the image branch in a single fused update:
\begin{equation}
\begin{aligned}
\hat{\mathbf{h}} &= \mathrm{SA}(\mathbf{h}),\quad
\mathbf{h}_{\mathrm{img}} = \mathrm{CA}(\hat{\mathbf{h}},\,\mathbf{e}_{\mathrm{clip}}),\quad
\mathbf{h}_{\mathrm{face}} = \mathrm{CA}(\hat{\mathbf{h}},\,\mathbf{E}_{\mathrm{face}}),\\
\tilde{\mathbf{h}}_{\mathrm{face}} &= \frac{\mathbf{h}_{\mathrm{face}}-\mu_{\mathrm{face}}}{\sigma_{\mathrm{face}}}\,\sigma_{\mathrm{img}} + \mu_{\mathrm{img}},\quad
\mathbf{h}_{\mathrm{out}} = \mathbf{h}_{\mathrm{img}} + \tilde{\mathbf{h}}_{\mathrm{face}}.
\end{aligned}
\label{eq:id_adapter_compact}
\end{equation}
Here \(\mathrm{SA}/\mathrm{CA}\) denote self‐/cross‐attention, \((\mu,\sigma)\) are the per‐token mean and standard deviation, and \(\mathbf{E}_{\mathrm{face}}\) the set of \(K\) identity tokens.  By matching the first and second moments of the face and image features, this adapter preserves identity information consistently across all frames.

\noindent\textbf{Human‐tracking masked reconstruction loss.}
Building upon StableAnimator’s face‐focused loss, we incorporate per‐person \emph{binary} masks for face and body regions.  
Original \(512\times512\) masks are downsampled via nearest‐neighbor interpolation to the latent resolution \(64\times64\).  
Given \(N\) individuals with binary masks  
$M^i_{\mathrm{face}},\,M^i_{\mathrm{body}}\in\{0,1\}^{1\times64\times64}$,
we optimize
\begin{equation}
\mathcal{L}_{\mathrm{rec}}
= \sum_{i=1}^{N}
  \mathbb{E}_{\epsilon\sim\mathcal{N}(0,1)}
  \Bigl\lVert
    (\mathbf{z}_{\mathrm{gt}} - \mathbf{z}_{\epsilon})
    \odot
    \bigl(1 + M^i_{\mathrm{body}} + 2\,M^i_{\mathrm{face}}\bigr)
  \Bigr\rVert_{2}^{2}.
\label{eq:human_masked_rec_loss}
\end{equation}
Here body masks have weight \(1\) and face masks weight \(2\), encouraging the model to focus capacity on identity‐critical regions while preserving overall reconstruction fidelity.

\subsection{MaskPoseAdapter}
\label{sec:maskpose}

Relying solely on pose keypoints (pose maps) makes it difficult to distinguish different individuals in multi-person scenarios; directly treating binary tracking masks as additional channels would compromise the translational invariance of the pose encoder. We therefore propose \textbf{MaskPoseAdapter}: first performing lightweight transformations on masks in the “pose feature space,” then injecting them into pose latents using a gated‐weighting strategy, and finally applying cross‐person soft‐attention to reorder per‐person importance. Fig.~\ref{fig:pipeline_overview} illustrates MaskPoseAdapter, which fuses independent pose streams and masks into a single pose–mask latent $\mathbf{F}\in\mathbb{R}^{B\times C\times64\times64}$.

\noindent\textbf{Per-person Pose Encoding.}  
For each person $i$, an independent \texttt{PoseNet0401} processes the RGB pose map $\mathbf{P}_i \in \mathbb{R}^{3 \times 512 \times 512}$. PoseNet consists of eight convolutional layers, expanding the channels from 3 to 128, followed by a $1 \times 1$ convolution, with weights shared across all persons. The output pose features are then scaled by a learnable factor $s$. The final output can be expressed as:
\begin{equation}
\mathbf{f}^{\text{pose}}_i = s \cdot \mathrm{Conv}_{1\times1}\bigl(\mathrm{PoseNet}(\mathbf{P}_i)\bigr) \;\in\; \mathbb{R}^{C\times64\times64},\; C=320,
\end{equation}

\noindent\textbf{Light Mask Processor.}  
Binary human/facial masks $\mathbf{M}_i \in \{0,1\}^{1 \times 512 \times 512}$ are processed through two $3 \times 3$ convolutional layers to produce a 3-channel feature map:
\begin{equation}
\mathbf{f}^{\text{mask}}_i = \psi(\mathbf{M}_i) \in \mathbb{R}^{3 \times 64 \times 64},
\end{equation}
which preserves contour information while avoiding mask features from dominating the pose features.

\noindent\textbf{Gate-based Fusion.}  
We apply two per-pixel gates to control how much of the pose and mask features to trust. These gates are implemented as convolutional layers followed by Sigmoid activations. The gate outputs are:
\begin{equation}
  w_i^{\mathrm{pose}} = \sigma\!\bigl(\gamma(\tilde{\mathbf{f}}^{\mathrm{pose}}_i)\bigr),\quad
  w_i^{\mathrm{mask}} = \sigma\!\bigl(\eta(\mathbf{f}^{\mathrm{mask}}_i)\bigr),
\end{equation}
where $\gamma$ and $\eta$ are each a \texttt{Conv→SiLU→Conv→Sigmoid} sequence. The gated features are then combined with a learnable weight $\lambda \approx 0.8$ as:
\begin{equation}
\tilde{\mathbf{f}}_i = \underbrace{\lambda\,w^{\mathrm{pose}}_i\odot\tilde{\mathbf{f}}^{\mathrm{pose}}_i}_{\text{ID-dominant}} 
\;+\; \underbrace{(1-\lambda)\,w^{\mathrm{mask}}_i\odot\mathbf{f}^{\mathrm{mask}}_i}_{\text{fine mask}},
\end{equation}
where $\tilde{\mathbf{f}}^{\mathrm{pose}}_i$ is the pose feature reduced to 3 channels for gating. A residual link is added to refine the fusion, where the coefficient \( \alpha_{\text{res}} \) controls the strength of the residual term:
\begin{equation}
  \mathbf{f}_i = \tilde{\mathbf{f}}_i + \alpha_{\text{res}}\,\bigl((1-\lambda)\,w_i^{\mathrm{mask}}\odot\mathbf{f}^{\mathrm{mask}}_i\bigr), 
  \quad \alpha_{\text{res}} = 0.5.
\end{equation}\label{equ:residual_link_alpha}

\noindent\textbf{Pose Enhancer.}  
The fusion output is passed through a lightweight \emph{PoseEnhancer} module consisting of a $3\times3$ convolution, followed by SiLU activation and BatchNorm, and a $1\times1$ convolution:
\begin{equation}
\mathbf{h}_i = \mathrm{PoseEnhancer}(\mathbf{f}_i).
\end{equation}
To further refine the pose features, a scaling factor $s_p = 1.5$ is applied to the raw features before final integration:
\begin{equation}
\mathbf{f}_i = s_p \cdot \mathbf{f}_i + (1-\alpha_{\text{res}}) \cdot \mathbf{h}_i.
\end{equation}

\noindent\textbf{LayerNorm and Attention.}  
Each of the enhanced pose features $\mathbf{f}_i$ is normalized per-channel using LayerNorm, resulting in $\bar{\mathbf{f}}_i$. The normalized features are concatenated along the channel dimension and processed through a lightweight attention mechanism consisting of three $1\times1$ convolution layers, each followed by BatchNorm and ReLU. This generates attention logits $\ell_i$ for each person:
\begin{equation}
\ell = \phi\bigl[\mathrm{LayerNorm}(\tilde{\mathbf{f}}_1), \dots, \mathrm{LayerNorm}(\tilde{\mathbf{f}}_N)\bigr] \in \mathbb{R}^{N\times64\times64}.
\end{equation}
These logits are normalized across the person dimension using a temperature-scaled softmax function:
\begin{equation}
\alpha_{\text{att}} = \mathrm{SoftmaxWithTemp}_{\tau}(\ell), \quad \mathrm{SoftmaxWithTemp}_{\tau}(x) = \mathrm{softmax}(x/\tau),
\end{equation}
where $\tau$ is a learnable temperature parameter.

\noindent\textbf{Cross-Person Integration.}  
We integrate the normalized features using both attention weighting and concatenation. First, we compute an attention-weighted sum of the features:
\begin{equation}
S = \sum_{i=1}^{N} \alpha_{\text{att},i} \odot \bar{\mathbf{f}}_i.
\end{equation}
Then, we pass the weighted sum through a $1\times1$ integration convolution to fuse the multi-person features into a final representation:
\begin{equation}
\mathbf{F}_{\mathrm{int}} = \mathrm{Conv}_{1\times1}(S).
\end{equation}
Finally
\begin{equation}
\mathbf{F} = 0.95 \cdot \mathbf{F}_{\mathrm{int}} + 0.05 \cdot \frac{1}{N} \sum_{i=1}^{N} \bar{\mathbf{f}}_i,
\end{equation}
where $\mathbf{F} \in \mathbb{R}^{C \times 64 \times 64}$ is reshaped to $(B,T,C,64,64)$ and injected into the UNet.

\subsection{MultiFace Encoder}\label{sec:fusionface}
For every mini-batch we receive $\mathbf{E}^{\mathrm{id}}\in\mathbb{R}^{N\times B\times D_{\mathrm{id}}}$ with $D_{\mathrm{id}}=512$ and $D_{\mathrm{clip}}=1024$, where the first axis enumerates the $N$ identities and the second the $B$ samples in the batch. Each sample also carries a length-1 CLIP embedding $\mathbf{e}^{\mathrm{clip}}\in\mathbb{R}^{B\times1\times D_{\mathrm{clip}}}$, which is used as key/value memory in all cross-attention steps.

\noindent\textbf{Stage I — Per-identity token projection.}
For identity $i\in\{1,\ldots,N\}$ and sample $b$ we transform the
ArcFace vector \(\mathbf{e}_{i,b}^{\mathrm{id}}\) with a two-layer MLP
(\texttt{Linear(512,1024) $\!\to$ GELU $\!\to$ Linear(1024,$KD$)})
and reshape it into \(K=4\) learnable tokens of width \(D=768\):
\begin{align}
\tilde{\mathbf{x}}_{i,b} &=
  \operatorname{MLP}_{2\times\mathrm{GELU}}
  (\mathbf{e}^{\mathrm{id}}_{i,b})
  \in\mathbb{R}^{K D},\\
\mathbf{t}^{(0)}_{i,b} &=
  \operatorname{LN}\!\bigl(
    \operatorname{reshape}_{K\times D}
    (\tilde{\mathbf{x}}_{i,b})\bigr)
  \in\mathbb{R}^{K\times D}.
\end{align}

\noindent\textbf{Stage II — FacePerceiver refinement.}
The $K$ latent tokens \(\mathbf{t}^{(0)}_{i,b}\) query a lightweight
\emph{FacePerceiver} with depth \(L_p=4\):
\begin{equation}
\mathbf{t}^{(\ell+1)}_{i,b}
=\mathbf{t}^{(\ell)}_{i,b}
+\mathrm{FFN}\!\Bigl(
   \mathbf{t}^{(\ell)}_{i,b}
  +\mathrm{CrossAttn}\bigl(\mathbf{t}^{(\ell)}_{i,b},
                           \mathbf{e}^{\mathrm{clip}}_b\bigr)\Bigr),
\quad\ell=0,\ldots,3.
\end{equation}
Queries originate from the latent tokens, whereas keys/values are the
concatenation of the projected CLIP embedding and the tokens
(cf.\ \texttt{PerceiverAttention} in the code).  
A residual shortcut controlled by the flags
\texttt{shortcut},\;\texttt{scale}\;(\(\lambda\)) reproduces the exact
behaviour of \texttt{MultiFace Encoder}:
\begin{equation}
\mathbf{t}_{i,b}=
\begin{cases}
\mathbf{t}^{(4)}_{i,b}, & \texttt{shortcut}=0,\\[4pt]
\mathbf{t}^{(0)}_{i,b}+\lambda\,\mathbf{t}^{(4)}_{i,b},
  & \texttt{shortcut}=1.
\end{cases}
\end{equation}

\noindent\textbf{Stage III — Multi-person concatenation.}
After processing all identities with \emph{shared} weights, the refined
tokens are stacked along the sequence axis:
\begin{equation}
\mathbf{T}_b
=\bigl[\mathbf{t}_{1,b};\,\mathbf{t}_{2,b};\,\dots;\,
       \mathbf{t}_{N,b}\bigr]
\in\mathbb{R}^{(N K)\times D},
\qquad
\mathbf{T}\in\mathbb{R}^{B\times N K\times D}\; \text{for the batch}.
\end{equation}
The UNet’s cross-attention layers can therefore read
\(\mathbf{T}\) directly, gaining \(N\!K\) extra tokens without any
architectural change.

\subsection{Data Curation Pipeline}\label{subsec:Data_Pipeline}
\begin{figure}[htbp]
    \centering
    \includegraphics[width=\linewidth]{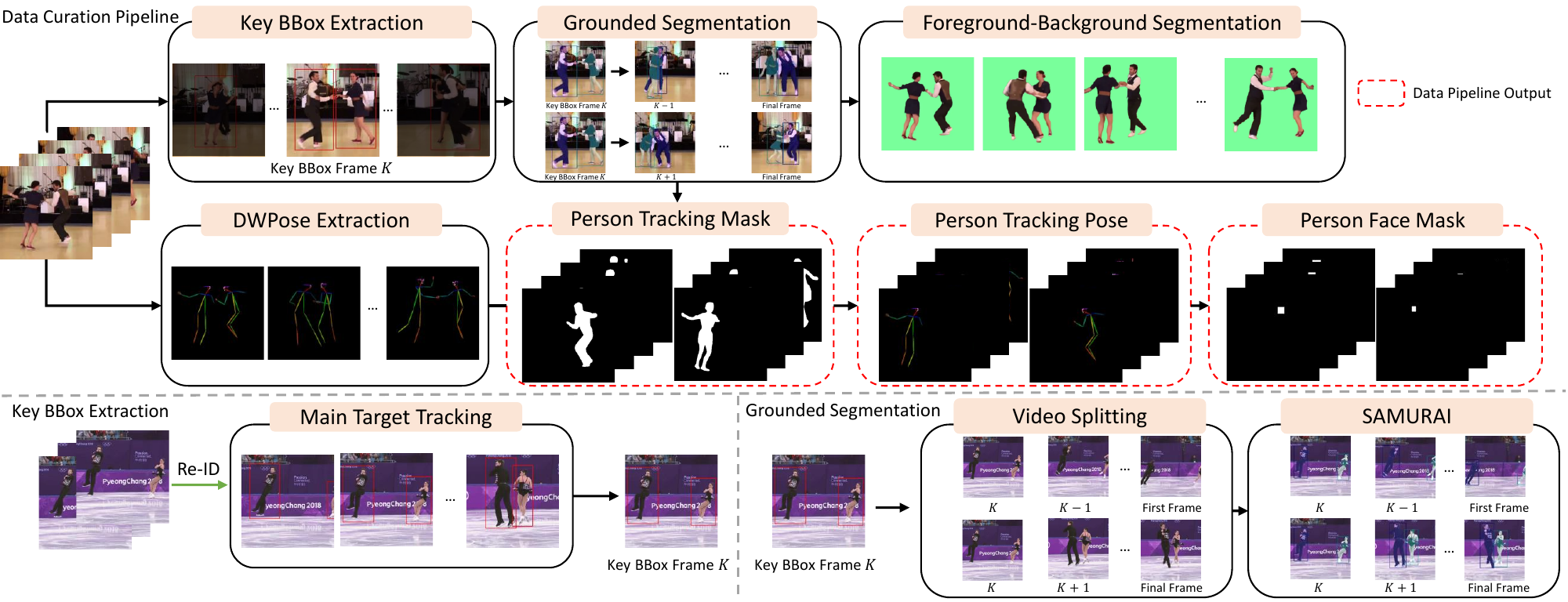}
    \caption{Data Curation Pipeline Overview. Our pipeline processes raw videos through human tracking, mask generation with SAMURAI~\cite{samba}, pose estimation with DW-Pose~\cite{dwpose}, and alpha matting to produce per-person annotations.}
    \label{fig:data-pipeline}
\end{figure}

To address the lack of two-person interaction datasets with diverse identities, static backgrounds, and fixed cameras, we propose a comprehensive data curation pipeline that recovers poses and mask annotations from monocular RGB videos.
As shown in Fig.~\ref{fig:data-pipeline}, our pipeline segments videos into scenes, detects and tracks individuals using YOLOv8x~\cite{yolov8_ultralytics} and OSNet-based ReID~\cite{zhou2019osnet,zhou2021osnet}, and selects primary subjects based on coverage and consistency. We then generate high-quality per-person masks and 133-point pose annotations using SAMURAI~\cite{yang2024samurai}, DWPose~\cite{dwpose}, and MatAnyone~\cite{yang2025matanyone}, followed by automatic and manual filtering to ensure data quality. We aggregate a wide range of single- and two-person motion datasets—including TikTokDataset~\cite{tiktokdataset}, Champ~\cite{zhu2024champ}, DisPose~\cite{li2025dispose}, HumanVid~\cite{wang2024humanvid}, Swing Dance~\cite{swingdance}, Harmony4D~\cite{khirodkar2024harmony4d}, CHI3D~\cite{CHI3D}, Beyond Talking~\cite{sun2025beyond}, and our newly collected \texttt{PairFS-4K}—to maximize identity diversity and interaction types. \texttt{PairFS-4K}, comprising 4.8K figure skating segments and over 7,000 unique identities, is the first large-scale two-person figure skating video dataset. All datasets are summarized in Tab.~\ref{tab:datasets-summary}, providing a rich foundation for controllable human interaction video generation in real-world scenarios.
More details of the Data Curation Pipeline can be found in Sec.~\ref{apeendix_datapipe}.
For specifics on the collection and processing of \texttt{PairFS-4K}, please refer to Sec.~\ref{PairFS_data_process}.

\begin{table}[htbp]
    \centering
    \caption{Summary of datasets used in DanceTogether training. $^*$Static competition background; $^\dagger$Static laboratory background; $^\ddagger$Multi-view setup.}
    \resizebox{\linewidth}{!}{
    \begin{tabular}{lccccccc}
    \toprule
    Dataset & Type & Action & IDs & Total & Avg. & Scene & Camera \\
    \midrule
    TikTokDataset~\cite{tiktokdataset} & Single & Dance & 332 & 1.03 hrs & 11 s & Static & Fixed \\
    Champ~\cite{zhu2024champ} & Single & Dance & 832 & 9.73 hrs & 42 s & Static & Fixed \\
    DisPose~\cite{li2025dispose} & Single & Dance & 8,636 & 38.12 hrs & 11 s & Static & Fixed \\
    HumanVid~\cite{wang2024humanvid} & Single & Dance & 16,310 & 89.89 hrs & 17 s & Dynamic & Moving \\
    \midrule
    Hi4D~\cite{yin2023hi4d} & Double & Interact & 40 & 0.10 hrs & 3.6 s & Static$^\dagger$ & Fixed$^\ddagger$ \\
    Harmony4D~\cite{khirodkar2024harmony4d} & Double & Interact & 24 & 0.58 hrs & 12 s & Static$^\dagger$ & Fixed$^\ddagger$ \\
    CHI3D~\cite{CHI3D} & Double & Interact & 6 & 1.75 hrs & 4 s & Static$^\dagger$ & Fixed$^\ddagger$ \\
    Swing Dance~\cite{swingdance} & Double & Dance & 1,356 & 23.36 hrs & 122 s & Static$^*$ & Moving \\
    HoCo~\cite{sun2025beyond} & Double & Talking Head & 26 & 45 hrs & 7 s & Static$^\dagger$ & Fixed \\
    \midrule
    \textbf{PairFS-4K} & Double & Figure Skating & 7,273 & 26.87 hrs & 20 s & Static$^*$ & Moving \\
    \textbf{HumanRob-300} & Single & Robot Interact & 336 & 0.83 hrs & 9 s & Dynamic & Moving \\
    \textbf{DanceTogEval-100} & Double & Interact \& Dance & 200 & 0.54 hrs & 20 s & Static & Fixed \\
    \bottomrule
    \end{tabular}}
    \label{tab:datasets-summary}
\end{table}

\subsection{TogetherVideoBench Benchmark}
We introduce \textbf{TogetherVideoBench}, a comprehensive benchmark for controllable multi-person video generation, which systematically evaluates three orthogonal tracks: \emph{Identity-Consistency}, \emph{Interaction-Coherence}, and \emph{Video Quality}. Please refer to details Sec.~\ref{our_benchmark} in the Appendix.

\section{Results}
\label{sec:results}
\subsection{Experimental Setup}
\label{subsec:Experimental_Setup}
We collect several publicly available video datasets, as detailed in Section~\ref{DatasetCollection}.
We utilize DWPose~\cite{dwpose} and ArcFace~\cite{deng2019arcface} to extract skeletal poses and facial embeddings/masks.
To evaluate the robustness of our model, we conduct experiments on DanceTogEval-100, a curated set of 100 previously unseen two-person interaction videos from the internet.
Following recent advances in animation generation~\cite{tu2024stableanimator}, we initialize our U-Net, PoseNet, and Face Encoder with the pre-trained weights from StableAnimator~\cite{blattmann2023stable}, then further train them on large-scale single-person datasets~\cite{tu2024stableanimator,li2025dispose,zhu2024champ,wang2024humanvid}.
We subsequently transfer the pre-trained weights to our proposed MaskPoseAdapter and MultiFace Encoder, and perform full fine-tuning using multi-person datasets—including our proposed PairFS dataset~\cite{swingdance,sun2025beyond,khirodkar2024harmony4d,CHI3D}.
Our model is trained for 20 epochs on 8 NVIDIA A100 80G GPUs, with a batch size of 1 per GPU and a learning rate set to $1e-5$.
\textbf{For ablation study, please refer to Sec.~\ref{append_ablation_study}.}

\subsection{Baselines}
\begin{figure}[h]
    \centering
    \includegraphics[width=\linewidth]{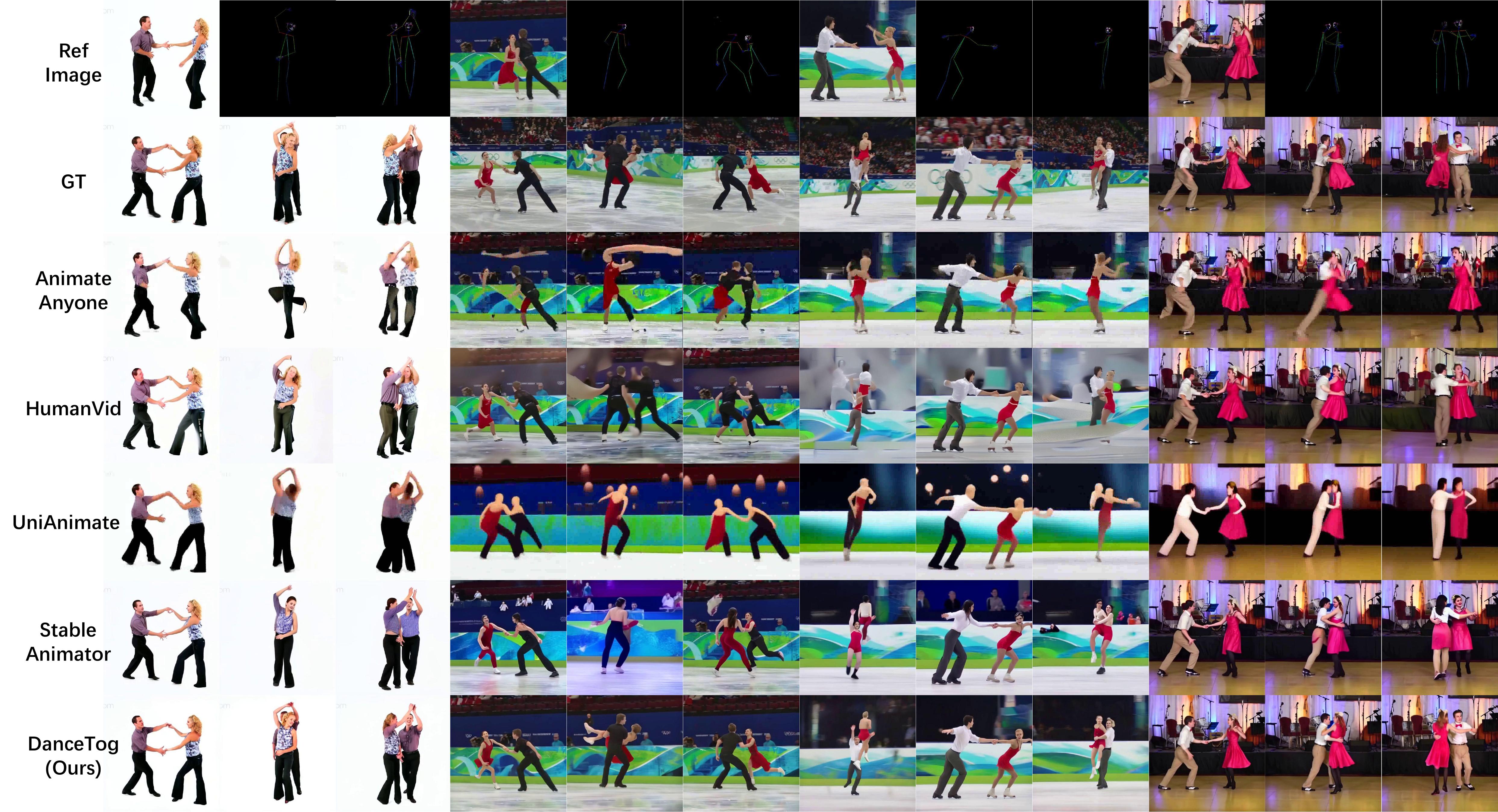}
    \caption{
    The RGB image in the “Ref Image” row is the input reference frame, and the two pose maps in that row correspond to the inference results shown immediately below.
    All baselines exhibit severe identity drift, loss of interaction details, or even missing subjects when dealing with position exchanges and complex interactive poses.
    For additional qualitative results, please refer to Appendix Fig.~\ref{fig:animation_result} and Fig.~\ref{fig:animation_result2}.
}
    \label{fig:comp-baseline}
\end{figure}

We compare our approach with state-of-the-art pose-conditioned human video generation models, including Animate Anyone~\cite{hu2024animate}, Champ~\cite{zhu2024champ}, MimicMotion~\cite{mimicmotion2024}, HumanVid~\cite{wang2024humanvid}, UniAnimate~\cite{wang2025unianimate}, UniAnimate-DiT~\cite{wang2025unianimateDiT}, DisPose~\cite{li2025dispose}, and StableAnimator~\cite{tu2024stableanimator}.
In particular, we fine-tune StableAnimator for 40 epochs on the dual-person dancing subset from the Swing Dance dataset~\cite{swingdance}, and include this fine-tuned variant as a new baseline in our evaluation.
Fig.~\ref{fig:comp-baseline} compares our proposed \emph{DanceTogether} with four strong baselines -- Animate Anyone~\cite{hu2024animate}, HumanVid~\cite{wang2024humanvid}, UniAnimate~\cite{wang2025unianimate}, and StableAnimator~\cite{tu2024stableanimator} -- all of which achieve relatively high scores in the quantitative evaluation. Additional comparisons, including more baselines and dual-person interaction examples, are provided in the appendix Sec.~\ref{MoreResults}.

\subsection{Quantitative Results}\label{sec:quantitative}

\noindent\textbf{Track~1: Identity--Consistency.}
Table~\ref{tab:combined_comparison} reports multiple‐object-tracking (MOT) scores on \textit{DanceTogEval-100}.  
Across all eight published baselines, StableAnimator fine-tuned on SwingDance (\textit{StableAnimator\,+$Data_{\text{swing}}$}) is the previous best performer, reaching \num{71.35} HOTA and \num{82.53}~IDF1.  
\emph{DanceTogether} markedly exceeds this strong baseline on \emph{every single metric}: with full training data it lifts HOTA from 71.35 to 81.79 (+\SI{10.44}{\percent}) and IDF1 from 82.53 to 87.73 (+\SI{6.3}{\percent}), while pushing AssA to \num{86.69}.  
Adding the proposed \texttt{PairFS-4K} dataset provides a further gain, culminating in \num{83.94} HOTA, \num{89.59}~IDF1, and a \num{79.80} MOTA.  
These results establish a new state of the art for long-range identity preservation under frequent occlusions and position exchanges.

\definecolor{shade1}{RGB}{255,140,0}   
\definecolor{shade2}{RGB}{255,165,70}  
\definecolor{shade3}{RGB}{255,190,120} 
\definecolor{shade4}{RGB}{255,210,150}  
\definecolor{shade5}{RGB}{255,225,180} 
\definecolor{shade6}{RGB}{255,240,230} 

\begin{table}[htbp]
  \centering
  \caption{Multiple Object Tracking results on \texttt{TogetherVideoBench}. $^*$ Negative values occur because the sum of false positives (FP) and false negatives (FN) exceeds the number of ground truth objects. This happens when the frames only contain a single person.}
  \resizebox{\linewidth}{!}{%
  \begin{tabular}{l|l|ccc|cc|c}
  \toprule
  \multirow{2}{*}{Method} & \multirow{2}{*}{Venue} &
  \multicolumn{3}{c|}{\textbf{HOTA family}} &
  \multicolumn{2}{c|}{\textbf{CLEAR/MOTA family}} &
  \textbf{Identity}\\
  \cmidrule(lr){3-5}\cmidrule(lr){6-7}\cmidrule(l){8-8}
   & & HOTA$\uparrow$ & DetA$\uparrow$ & AssA$\uparrow$ & MOTA$\uparrow$ & MOTP$\uparrow$ & IDF1$\uparrow$ \\
  \midrule
Animate Anyone~\cite{hu2024animate} & CVPR 2024 &
  41.26 & 39.99 & 43.21 & 26.67 & 75.73 & 51.54 \\
Champ~\cite{zhu2024champ} & ECCV 2024 &
  19.32 & 14.78 & 26.32 & -19.54$^*$ & 67.92 & 17.84 \\
MimicMotion~\cite{mimicmotion2024} & Arxiv 2024 &
  21.14 & 16.06 & 30.50 & -55.77$^*$ & 62.13 & 15.24 \\
HumanVid~\cite{wang2024humanvid} & NeurIPS 2024 &
  \cellcolor{shade6}56.12 & \cellcolor{shade6}58.89 & \cellcolor{shade6}53.69 & \cellcolor{shade6}58.86 & \cellcolor{shade6}84.20 & \cellcolor{shade6}68.84 \\
UniAnimate~\cite{wang2025unianimate} & SCIS 2025 &
  48.43 & 46.71 & 50.69 & 42.33 & 80.74 & 59.58 \\
UniAnimate-DiT~\cite{wang2025unianimateDiT} & Arxiv 2025 &
  35.02 & 31.15 & 40.65 & 10.66 & 77.99 & 39.51 \\
DisPose~\cite{li2025dispose} & ICLR 2025 &
  20.68 & 15.91 & 29.47 & -52.49$^*$ & 62.00 & 15.42 \\
StableAnimator~\cite{tu2024stableanimator} & CVPR 2025 &
  \cellcolor{shade5}67.75 & \cellcolor{shade5}67.91 & \cellcolor{shade5}67.70 & \cellcolor{shade5}69.62 & \cellcolor{shade5}87.67 & \cellcolor{shade5}79.37 \\
StableAnimator w. $Data_{swing}$ & CVPR 2025 &
  \cellcolor{shade4}71.35 & \cellcolor{shade4}70.91 & \cellcolor{shade4}71.89 & \cellcolor{shade3}73.89 & \cellcolor{shade4}88.22 & \cellcolor{shade4}82.53 \\
\midrule
\textbf{DanceTog w. $Data_{swing}$} & \textbf{--} &
  \cellcolor{shade3}80.26 & \cellcolor{shade3}74.44 & \cellcolor{shade3}86.57 & \cellcolor{shade4}73.68 & \cellcolor{shade3}95.45 & \cellcolor{shade3}86.28 \\
\textbf{DanceTog w. $Data_{full}$} & \textbf{--} &
  \cellcolor{shade2}81.79 & \cellcolor{shade2}77.19 & \cellcolor{shade2}86.69 & \cellcolor{shade2}77.04 & \cellcolor{shade1}95.69 & \cellcolor{shade2}87.73 \\
\textbf{DanceTog w. $Data_{full}+Data_{PairFS}$} & \textbf{--} &
  \cellcolor{shade1}83.94 & \cellcolor{shade1}79.48 & \cellcolor{shade1}88.68 & \cellcolor{shade1}79.80 & \cellcolor{shade2}95.49 & \cellcolor{shade1}89.59 \\
\bottomrule
  \end{tabular}}
  \label{tab:combined_comparison}
\end{table}

\noindent\textbf{Track~2: Interaction--Coherence.}
Table~\ref{tab:interaction_coherence} evaluates how faithfully each method follows the target motion and how smoothly the interaction unfolds.  
Our model slashes $\mathrm{MPJPE}_{2\text{D}}$ by \SI{68}{\percent} relative to the top baseline (from \num{1555}\,px to \num{492}\,px) and attains the highest OKS (\num{0.83}) and PoseSSIM (\num{0.93}).  
At the same time, \emph{DanceTogether} records the lowest motion-discontinuity scores—SmoothRMS~\num{0.83e6} and TimeDyn$_{\text{RMSE}}$~\num{1.59e4}—indicating physically plausible, temporally consistent choreography. 
Champ achieves high scores on SmoothRMS and TimeDyn$_{\text{RMSE}}$ due to its use of estimated SMPL as guidance, which incorporates smoothing methods in the process of generating SMPL sequences. These two metrics only compare the motion continuity of each individual person without applying weights to the pair. Champ's inference results typically contain only a single person; for qualitative comparison results, please refer to Sec.~\ref{MoreResults}.
The FVMD is halved compared with \textit{StableAnimator} (0.54 vs.\ 1.00), further corroborating superior interaction quality.

\definecolor{shade1}{RGB}{255,140,0}   
\definecolor{shade2}{RGB}{255,165,70}  
\definecolor{shade3}{RGB}{255,190,120} 
\definecolor{shade4}{RGB}{255,210,150}  
\definecolor{shade5}{RGB}{255,225,180} 
\definecolor{shade6}{RGB}{255,240,230} 
\begin{table}[htbp]
\centering
\caption{Comparison of models across interaction coherence metrics. }
\resizebox{\linewidth}{!}{%
\begin{tabular}{l|cccccc}
\toprule
\multirow{2}{*}{Method} & MPJPE$_{2D}$$\downarrow$ & OKS$\uparrow$ & PoseSSIM$\uparrow$ & SmoothRMS$\downarrow$ & TimeDyn$_{RMSE}$$\downarrow$ & FVMD$\downarrow$ \\
& & & & ($\times 10^6$) & ($\times 10^4$) & ($\times 10^5$) \\
\midrule
Animate Anyone & 3255.07 & 0.27 & 0.67 & 1.26 & 2.43 & 1.87 \\
Champ & 4117.88 & 0.06 & \cellcolor{shade6}0.78 & \cellcolor{shade1}0.78 & \cellcolor{shade2}1.60 & \cellcolor{shade5}0.90 \\
MimicMotion & 5542.99 & 0.09 & 0.74 & 1.02 & 1.94 & 1.15 \\
HumanVid & 3480.74 & \cellcolor{shade6}0.48 & \cellcolor{shade6}0.78 & 1.09 & 2.10 & 1.11 \\
UniAnimate & 2286.26 & 0.37 & 0.72 & 1.24 & 2.36 & 2.13 \\
UniAnimate-DiT & \cellcolor{shade6}2184.81 & 0.22 & 0.71 & 1.53 & 2.92 & 3.72 \\
DisPose & 2791.60 & 0.08 & 0.73 & 1.07 & 2.04 & 1.36 \\
StableAnimator & \cellcolor{shade5}1571.50 & \cellcolor{shade5}0.63 & \cellcolor{shade5}0.82 & \cellcolor{shade6}0.96 & \cellcolor{shade6}1.84 & \cellcolor{shade6}1.00 \\
StableAnimator w. $Data_{swing}$ & \cellcolor{shade4}1555.16 & \cellcolor{shade4}0.70 & \cellcolor{shade4}0.84 & \cellcolor{shade5}0.89 & \cellcolor{shade5}1.72 & \cellcolor{shade4}0.77 \\
\midrule
\textbf{DanceTog w. $Data_{swing}$} & \cellcolor{shade3}858.99 & \cellcolor{shade3}0.75 & \cellcolor{shade3}0.88 & \cellcolor{shade3}0.84 & \cellcolor{shade3}1.62 & \cellcolor{shade1}0.51 \\
\textbf{DanceTog w. $Data_{full}$} & \cellcolor{shade2}557.60 & \cellcolor{shade2}0.81 & \cellcolor{shade2}0.92 & \cellcolor{shade4}0.85 & \cellcolor{shade4}1.64 & \cellcolor{shade3}0.66 \\
\textbf{DanceTog w. $Data_{full}+Data_{PairFS}$} & \cellcolor{shade1}492.24 & \cellcolor{shade1}0.83 & \cellcolor{shade1}0.93 & \cellcolor{shade2}0.83 & \cellcolor{shade1}1.59 & \cellcolor{shade2}0.54 \\
\bottomrule
\end{tabular}}
\label{tab:interaction_coherence}
\end{table}

\noindent\textbf{Track~3: Video Quality.}
Tables~\ref{tab:full_frame} and \ref{tab:masked_region} present full-frame and mask-aware appearance metrics.  
Benefiting from dense identity–action binding and the high-diversity PairFS-4K corpus, \emph{DanceTogether} delivers the best perceptual fidelity in both settings.  
In full-frame evaluation it attains the lowest FVD (\num{76.3}) and FID (\num{75.1}), alongside the highest CLIP score (\num{0.95}) and ST-SSIM (\num{0.70}).  
Within the human-masked regions—the areas most sensitive to identity drift—mask-aware FVD plunges from 29.0 to \textbf{17.1}, and C-FID shrinks from 12.5 to \textbf{7.9}, highlighting crisp texture reproduction and identity accuracy.  
Notably, these improvements are achieved without sacrificing low-level reconstruction fidelity: L1 and LPIPS fall in tandem, while PSNR and SSIM increase.

\definecolor{shade1}{RGB}{255,140,0}   
\definecolor{shade2}{RGB}{255,165,70}  
\definecolor{shade3}{RGB}{255,190,120} 
\definecolor{shade4}{RGB}{255,210,150}  
\definecolor{shade5}{RGB}{255,225,180} 
\definecolor{shade6}{RGB}{255,240,230} 
\begin{table}[htbp]
  \centering
  \caption{Comparison of models using \textbf{Full Frame} evaluation metrics.}
  \resizebox{\linewidth}{!}{%
  \begin{tabular}{l|ccccccccccc}
  \toprule
  Method & L1$\downarrow$ & PSNR$\uparrow$ & SSIM$\uparrow$ & LPIPS$\downarrow$ & DISTS$\downarrow$ & CLIP$\uparrow$ & ST-SSIM$\uparrow$ & GMSD-T$\downarrow$ & FVD$\downarrow$ & FID$\downarrow$ & C-FID$\downarrow$ \\
  \midrule
AnimateAnyone               & \cellcolor{shade6}37.32 & 13.23               & 0.49               & 0.56               & 0.27               & 0.91               & 0.54               & 0.42               & 108.2              & 118.1              & 27.7               \\
Champ                       & 43.70                  & 11.93               & 0.49               & 0.56               & 0.29               & 0.91               & 0.39               & \cellcolor{shade2}0.36 & 125.7              & 114.6              & 25.6               \\
MimicMotion                 & 52.08                  & 11.04               & 0.47               & 0.58               & 0.32               & 0.91               & 0.37               & \cellcolor{shade4}0.39               & 121.0              & 116.6              & 26.2               \\
HumanVid                    & 38.93                  & \cellcolor{shade6}13.67 & 0.52               & \cellcolor{shade6}0.50 & \cellcolor{shade6}0.26 & \cellcolor{shade6}0.93 & 0.53               & \cellcolor{shade1}0.35 & \cellcolor{shade6}97.2  & \cellcolor{shade6}90.2  & \cellcolor{shade6}18.6  \\
UniAnimate                  & 37.95                  & 13.62               & \cellcolor{shade6}0.55 & 0.53               & 0.29               & 0.89               & \cellcolor{shade6}0.61 & 0.42               & 132.0              & 151.2              & 42.8               \\
UniAnimate-DiT              & 43.11                  & 12.34               & 0.50               & 0.53               & 0.28               &     0.92 & 0.45               & 0.42               & 111.9              & 100.3              & 20.8  \\
DisPose                     & 42.52                  & 12.28               &    0.54 &  0.54               & 0.31               & 0.91               & 0.41               & \cellcolor{shade4}0.39 & 127.4              & 127.9              & 31.0               \\
StableAnimator              & \cellcolor{shade5}33.44 & \cellcolor{shade5}14.60 & \cellcolor{shade5}0.57 & \cellcolor{shade4}0.44 & \cellcolor{shade5}0.24 & \cellcolor{shade3}0.94 & \cellcolor{shade5}0.66 & 0.40               & \cellcolor{shade5}85.7  & \cellcolor{shade5}84.1  & \cellcolor{shade5}18.1  \\
StableAnimator w.\,$Data_{swing}$ & \cellcolor{shade3}30.31 & \cellcolor{shade3}15.27 & \cellcolor{shade3}0.60 & \cellcolor{shade1}0.42 & \cellcolor{shade1}0.22 & \cellcolor{shade3}0.94 & \cellcolor{shade3}0.69 & 0.42               & \cellcolor{shade3}78.8  & \cellcolor{shade3}79.3  & \cellcolor{shade4}16.1  \\
\midrule
\textbf{DanceTog w.\,$Data_{swing}$}        & \cellcolor{shade4}32.62 & \cellcolor{shade4}15.12 & \cellcolor{shade4}0.59 & \cellcolor{shade4}0.44 & \cellcolor{shade4}0.23 & \cellcolor{shade3}0.94 & \cellcolor{shade4}0.68 & \cellcolor{shade3}0.38 & \cellcolor{shade4}79.3  & \cellcolor{shade4}82.1  & \cellcolor{shade3}14.7  \\
\textbf{DanceTog w.\,$Data_{full}$}         & \cellcolor{shade2}29.94 & \cellcolor{shade2}15.81 & \cellcolor{shade1}0.61 & \cellcolor{shade1}0.42 & \cellcolor{shade1}0.22 & \cellcolor{shade1}0.95 & \cellcolor{shade1}0.70 & \cellcolor{shade4}0.39 & \cellcolor{shade2}76.9  & \cellcolor{shade2}77.6  & \cellcolor{shade2}13.1  \\
\textbf{DanceTog w.\,$Data_{full}+Data_{PairFS}$} & \cellcolor{shade1}29.52 & \cellcolor{shade1}15.85 & \cellcolor{shade1}0.61 & \cellcolor{shade1}0.42 & \cellcolor{shade1}0.22 & \cellcolor{shade1}0.95 & \cellcolor{shade1}0.70 & \cellcolor{shade4}0.39 & \cellcolor{shade1}76.3  & \cellcolor{shade1}75.1  & \cellcolor{shade1}12.6  \\
  \bottomrule
  \end{tabular}}
  \label{tab:full_frame}
\end{table}

\begin{table}[htbp]
  \centering
  \caption{Comparison of models using \textbf{Human Masked Region} evaluation metrics.}
  \resizebox{\linewidth}{!}{%
  \begin{tabular}{l|ccccccccccc}
  \toprule
  Method & L1$\downarrow$ & PSNR$\uparrow$ & SSIM$\uparrow$ & LPIPS$\downarrow$ & DISTS$\downarrow$ & CLIP$\uparrow$ & ST-SSIM$\uparrow$ & GMSD-T$\downarrow$ & FVD$\downarrow$ & FID$\downarrow$ & C-FID$\downarrow$ \\
  \midrule
AnimateAnyone               & 59.92                 & 10.45               & 0.92               & 0.06               & 0.13               & 0.92               & 0.70               & 0.18               & 44.8               & 101.4              & 19.2               \\
Champ                       & 83.35                 & 8.36                & 0.92               & 0.07               & 0.17               & 0.90               & 0.58               & 0.17               & 69.2               & 178.7              & 34.2               \\
MimicMotion                 & 77.02                 & 8.75                & 0.92               & 0.07               & 0.16               & 0.90               & 0.56               & 0.17               & 65.5               & 180.9              & 33.9               \\
HumanVid                    & \cellcolor{shade5}47.97 & \cellcolor{shade5}12.13 & \cellcolor{shade4}0.93 & \cellcolor{shade5}0.05 & \cellcolor{shade5}0.12 & \cellcolor{shade5}0.93 & \cellcolor{shade5}0.76 & \cellcolor{shade5}0.15 & \cellcolor{shade5}34.9  & \cellcolor{shade6}72.4   & \cellcolor{shade5}14.2  \\
UniAnimate                  & 56.34                 & 11.05               & 0.92               & 0.06               & 0.13               & 0.92               & 0.70               & 0.17               & 45.0               & 109.8              & 21.4               \\
UniAnimate-DiT              & 64.48                 & 9.89                & 0.91               & 0.06               & 0.14               & 0.90               & 0.68               & 0.18               & 51.4               & 119.6              & 21.5               \\
DisPose                     & 76.75                 & 8.93                & 0.92               & 0.07               & 0.16               & 0.90               & 0.60               & 0.17               & 64.7               & 196.0              & 36.4               \\
StableAnimator              & \cellcolor{shade6}48.51 & \cellcolor{shade6}12.00 & \cellcolor{shade4}0.93 & \cellcolor{shade5}0.05 & \cellcolor{shade5}0.12 & \cellcolor{shade5}0.93 & \cellcolor{shade6}0.75 & \cellcolor{shade5}0.15 & \cellcolor{shade6}38.4  & \cellcolor{shade5}71.8   & \cellcolor{shade6}15.7  \\
StableAnimator w.\,$Data_{swing}$ & \cellcolor{shade4}41.41 & \cellcolor{shade4}13.06 & \cellcolor{shade4}0.93 & \cellcolor{shade4}0.04 & \cellcolor{shade4}0.11 & \cellcolor{shade2}0.94 & \cellcolor{shade4}0.80 & \cellcolor{shade1}0.14 & \cellcolor{shade4}29.0  & \cellcolor{shade4}66.7   & \cellcolor{shade4}12.5  \\
\midrule
\textbf{DanceTog w.\,$Data_{swing}$}        & \cellcolor{shade3}34.49 & \cellcolor{shade3}14.76 & \cellcolor{shade1}0.94 & \cellcolor{shade1}0.03 & \cellcolor{shade2}0.09 & \cellcolor{shade2}0.94 & \cellcolor{shade2}0.85 & \cellcolor{shade1}0.14 & \cellcolor{shade3}21.5  & \cellcolor{shade3}57.5   & \cellcolor{shade3}9.5   \\
\textbf{DanceTog w.\,$Data_{full}$}         & \cellcolor{shade2}32.80 & \cellcolor{shade2}15.15 & \cellcolor{shade1}0.94 & \cellcolor{shade1}0.03 & \cellcolor{shade2}0.09 & \cellcolor{shade2}0.94 & \cellcolor{shade2}0.85 & \cellcolor{shade1}0.14 & \cellcolor{shade2}20.6  & \cellcolor{shade2}56.1   & \cellcolor{shade2}8.9   \\
\textbf{DanceTog w.\,$Data_{full}+Data_{PairFS}$} & \cellcolor{shade1}30.14 & \cellcolor{shade1}15.82 & \cellcolor{shade1}0.94 & \cellcolor{shade1}0.03 & \cellcolor{shade1}0.08 & \cellcolor{shade1}0.95 & \cellcolor{shade1}0.87 & \cellcolor{shade1}0.14 & \cellcolor{shade1}17.1  & \cellcolor{shade1}48.0   & \cellcolor{shade1}7.9   \\
  \bottomrule
  \end{tabular}}
  \label{tab:masked_region}
\end{table}

\section{Conclusion}\label{sec:conclusion}

We present DanceTogether, the first end-to-end diffusion framework for generating long, photorealistic multi-actor videos from a single reference image and independent pose–mask streams, while strictly preserving each identity. Our method integrates a novel MaskPoseAdapter for persistent identity–action alignment and a MultiFace Encoder for compact appearance encoding. Trained on our newly curated multi-actor datasets and evaluated on a comprehensive benchmark, DanceTogether outperforms all existing pose-conditioned video generation models by a significant margin. It generalizes well across domains, as demonstrated by convincing human–robot interactions after minimal adaptation. This work marks a step forward toward compositionally controllable, identity-aware video synthesis, laying a foundation for future advances in digital content creation, simulation, and embodied AI.

{
\small
\bibliographystyle{ieee_fullname}
\bibliography{main}
}
\appendix

\section{Limitations}\label{Limitations}

While DanceTogether achieves state-of-the-art performance on two-person interaction benchmarks, it has several limitations. First, our framework is optimized for up to two actors; extending it to handle larger groups would incur substantial computational and memory overhead and may require hierarchical or factorized conditioning mechanisms. Second, the quality of generated videos depends heavily on the accuracy of the input pose and mask sequences—severe occlusions, fast motion blur, or failures in the underlying detectors (e.g., DWPose~\cite{dwpose}, SAMURAI~\cite{yang2024samurai}) can degrade identity preservation and interaction fidelity. Third, we assume a mostly static camera and relatively simple backgrounds; dynamic camera motion or highly cluttered scenes may introduce artifacts or identity confusion. Fourth, like most diffusion-based methods, DanceTogether is computationally intensive and incurs non-trivial latency, limiting real-time applications.

\section{Broader impacts}\label{Broader_impacts}

DanceTogether opens new possibilities for creative content production, digital avatar animation, and embodied-AI simulation by enabling controllable, identity-preserving multi-person video generation. It can accelerate workflows in film, game, and VR/AR industries, and provide high-fidelity training data for human-robot interaction research. However, the ability to generate realistic multi-person videos also raises potential misuse risks—such as deepfake creation, identity impersonation, and privacy infringements. Our large-scale datasets (PairFS-4K, HumanRob-300) may inadvertently encode demographic biases; we therefore recommend careful curation and bias analysis before deployment. To mitigate misuse, we plan to release public checkpoints with visible watermarks and to accompany the code and models with clear ethical guidelines and usage licenses. We believe that, with appropriate safeguards, DanceTogether can serve as a responsible tool for advancing both research and creative industries.

\section{Ablation Study}\label{append_ablation_study}
\subsection{Dataset ablation study}
Ablation study on the datasets have been compared in the main text in Tabs.~\ref{tab:combined_comparison}, ~\ref{tab:interaction_coherence}, ~\ref{tab:full_frame}, and ~\ref{tab:masked_region}. 
StableAnimator~\cite{tu2024stableanimator} fine-tuned for 40 epochs on the swing dance dataset~\cite{swingdance} (StableAnimator w. $Data_{swing}$) shows significant improvement over the original pre-trained weights provided by the authors, but still performs noticeably worse than DanceTog trained for 20 epochs on the same Swing dance dataset (\textbf{DanceTog w. $Data_{swing}$}). Using all training data except PairFS-4K (\textbf{DanceTog w. $Data_{full}$}) clearly performs better than the model trained only on the swing dance dataset (\textbf{DanceTog w. $Data_{swing}$}), but still underperforms compared to DanceTog trained on all data including PairFS-4K (denoted as \textbf{DanceTog w. $Data_{full}+Data_{PairFS}$}). 

\subsection{Ablation study on sub-modules and inputs}

Tab.~\ref{tab:abl-module} provides ablation results for the new model and multi-input approaches proposed in DanceTog.
Where, w/o mask input means not using a separate mask input during the input process.
w/o pose input means not using a separate pose input during the input process.
w/o MaskPoseAdapter means using the original PoseNet, i.e., using the poses of all people as condition inputs to the model.
w/o MultiFaceEncoder means using the original FaceEncoder, i.e., using the embedding of the largest face detected in the reference image as a condition input to the model.

\begin{table*}[htbp]
\centering
\small
\setlength{\tabcolsep}{4pt}
\renewcommand{\arraystretch}{1.15}
\resizebox{\linewidth}{!}{%
\begin{tabular}{l | ccc | cccc | cccc}
\toprule
\multirow{2}{*}{\textbf{Model Variant}} 
  & \multicolumn{3}{c|}{Track 1: Identity–Consistency} 
  & \multicolumn{4}{c|}{Track 2: Interaction–Coherence} 
  & \multicolumn{4}{c}{Track 3: Video Quality} \\  
\cmidrule(lr){2-4}\cmidrule(lr){5-8}\cmidrule(lr){9-12} 
  & HOTA$\uparrow$ & MOTA$\uparrow$ & IDF1$\uparrow$ 
  & MPJPE$_{2D}$$\downarrow$ & OKS$\uparrow$ & PoseSSIM$\uparrow$ & FVMD$\times10^5$$\downarrow$ 
  & PSNR$\uparrow$ & FVD$\downarrow$ & FID$\downarrow$ & C-FID$\downarrow$ \\ 
\midrule
w/o mask input                      
  & 33.63 & 15.48 & 42.49 & 1625.04 & 0.28 & 0.85 & 2.97 & 11.02 & 40.4 & 73.1 & 14.7 \\
w/o pose input                      
  & 81.48 & 74.23 & 86.38 & 1292.33 & 0.46 & 0.85 & 4.91 & 14.98 & 19.7 & 58.1 & 9.4  \\
w/o MaskPoseAdapter                 
  & 48.95 & 40.93 & 62.02 & 1692.55 & 0.48 & 0.79 & 3.80 & 11.19 & 41.3 & 72.0 & 14.2 \\
w/o MultiFaceEncoder                
  & 83.31 & 78.81 & 88.55 &  893.32 & 0.74 & 0.89 & 1.26 & 15.67 & 17.9 & 49.2 & 8.4  \\
\midrule
\textbf{DanceTog}              
  & 83.94 & 79.80 & 89.59 &  492.24 & 0.83 & 0.93 & 0.54 & 15.82 & 17.1 & 48.0 & 7.9  \\ 
\bottomrule
\end{tabular}
}
\caption{
Module ablation study.
}
\label{tab:abl-module}
\end{table*}

\begin{figure}[htbp]
    \centering
    \includegraphics[width=1\linewidth]{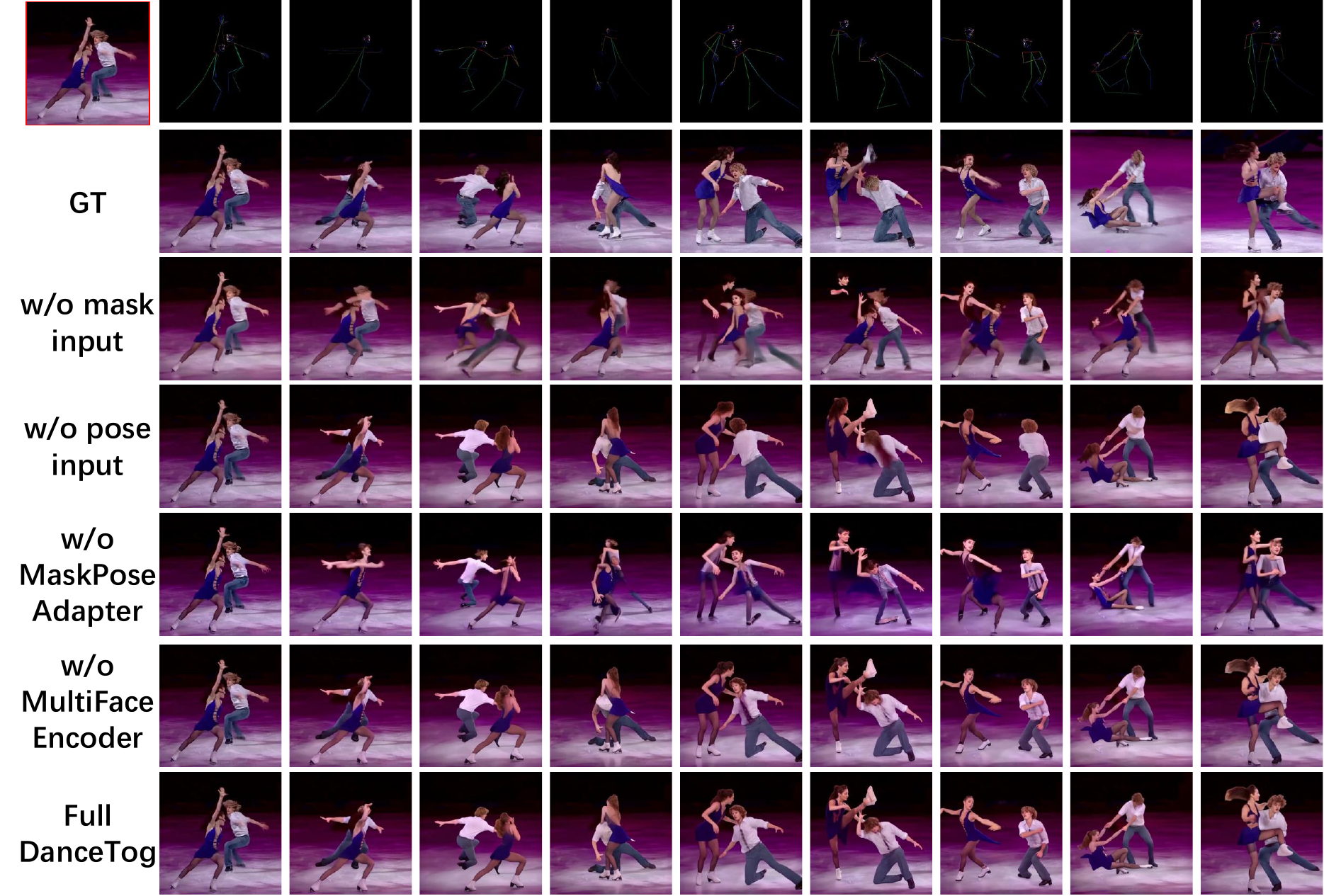}
    \caption{Ablation study animation results (1/2).}
    \label{fig:ablation_case01}
\end{figure}

\begin{figure}[htbp]
    \centering
    \includegraphics[width=1\linewidth]{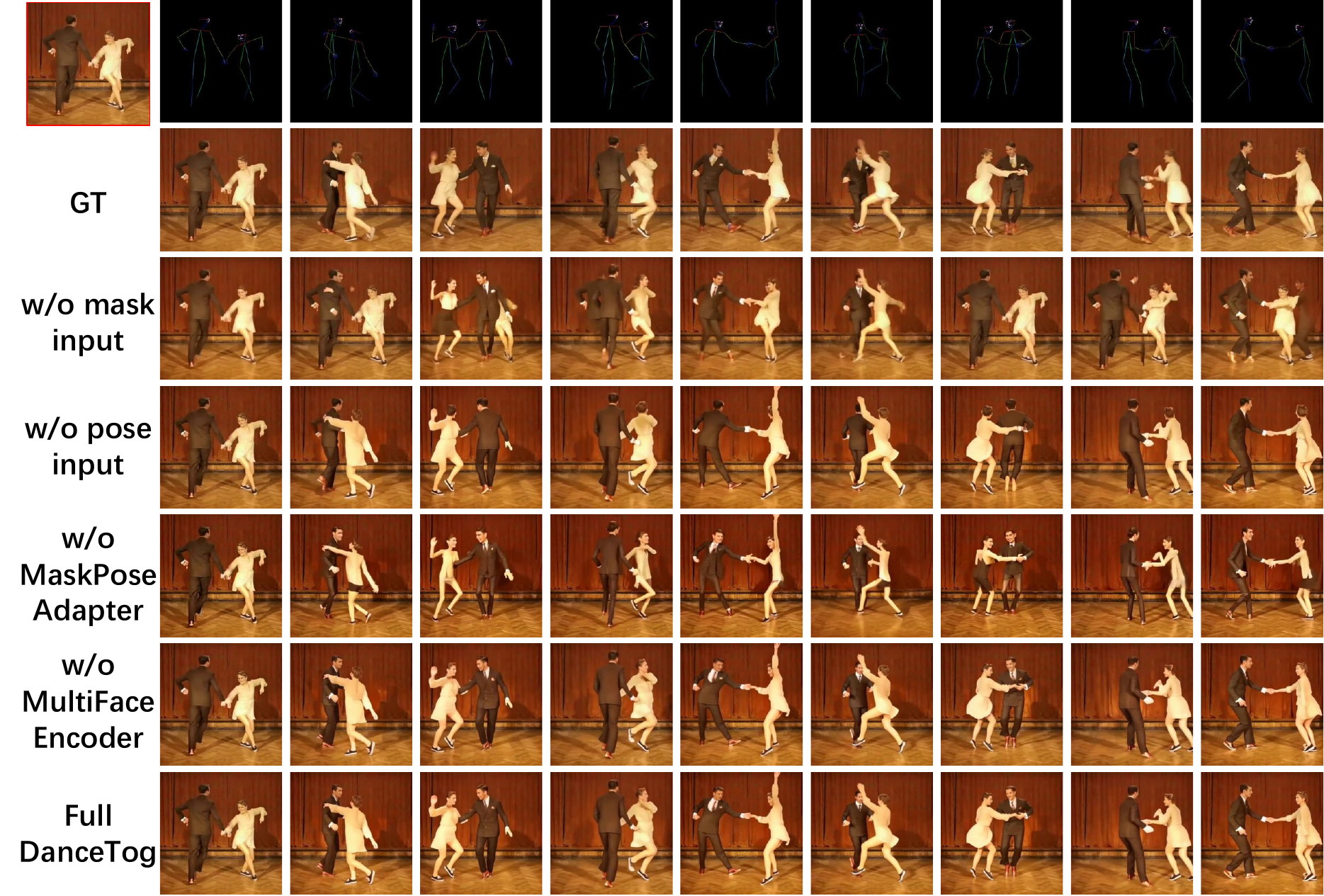}
    \caption{Ablation study animation results (2/2).}
    \label{fig:ablation_case02}
\end{figure}

Fig.~\ref{fig:ablation_case01} and Fig.~\ref{fig:ablation_case02} present qualitative comparisons of our ablation studies. 
DanceTogether is compatible with StableAnimator’s “Inference with HJB-based Face Optimization”~\cite{tu2024stableanimator}.
Since our task and test samples focus on full-body two-person interaction video generation rather than large-area face-mask talking heads or single-person half-body dance sequences, the benefit of HJB-based Face Optimization is less pronounced. 
In our tests, inference without HJB-based Face Optimization runs at approximately 0.8\,s/iteration, whereas enabling HJB-based Face Optimization reduces throughput to about 15\,s/iteration. Furthermore, our ablation study indicates that applying HJB-based Face Optimization does not significantly impact the quality of two-person interaction video generation. Consequently, all experiments reported in the main text for StableAnimator and DanceTog were performed without HJB-based Face Optimization.

\begin{figure}[htbp]
    \centering
    \includegraphics[width=1\linewidth]{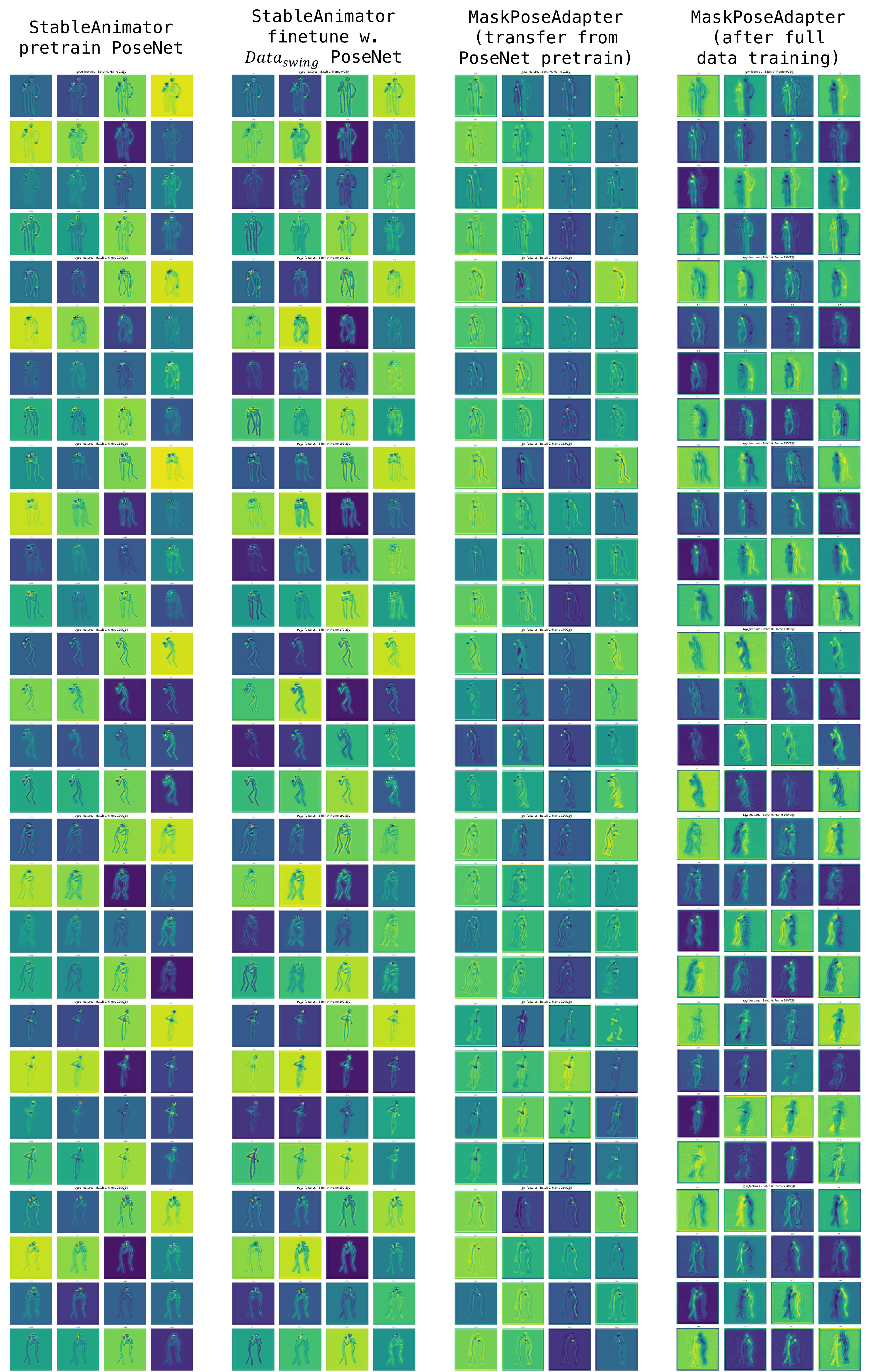}
    \caption{ 
    Comparison of PoseNet and MaskPoseAdapter outputs under identical frame inputs.
    }
    \label{fig:comp_posenet}
\end{figure}

\subsection{Comparison between PoseNet and MaskPoseAdapter}
Fig.~\ref{fig:comp_posenet} shows the feature maps obtained by the original PoseNet and our proposed MaskPoseAdapter from consecutive frames with the same input. It can be clearly observed that the output of MaskPoseAdapter strongly binds pose and mask information, enabling clear identification of which ID each pose corresponds to, and still providing sufficient mask information even when input poses are missing in some occluded frames.
In contrast, the original PoseNet's output makes it difficult to distinguish each individual pose, and pose features may be lost in occluded frames.

\begin{figure}[htbp]  
    \centering
    \begin{minipage}{\linewidth}
        \centering
        \includegraphics[width=\linewidth]{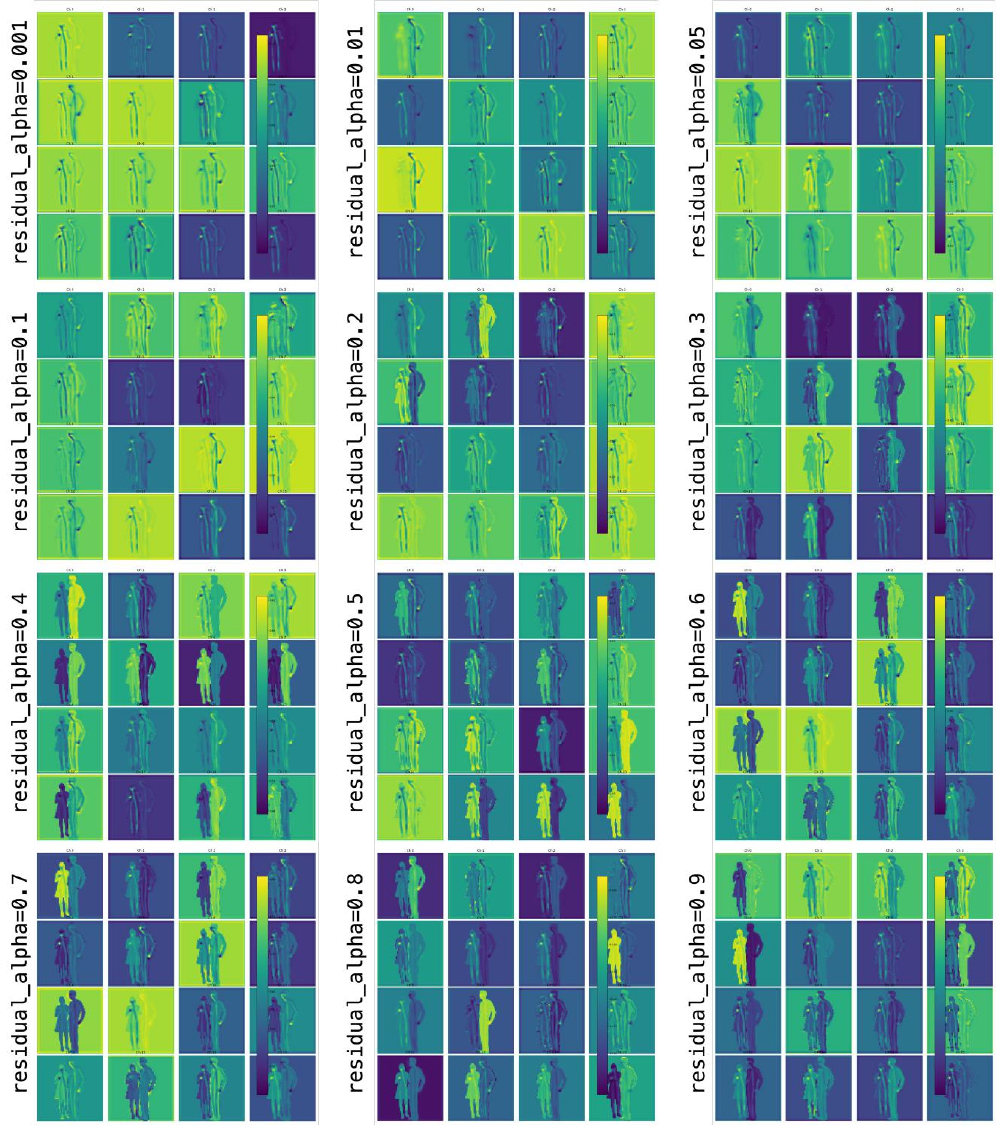}
        \caption{
        The effect of residual alpha on MaskPoseAdapter output.
        }
        \label{fig:residual_alpha}
    \end{minipage}
    
    \vspace{1em}  
    
    \begin{minipage}{\linewidth}
        \centering
        \includegraphics[width=\linewidth]{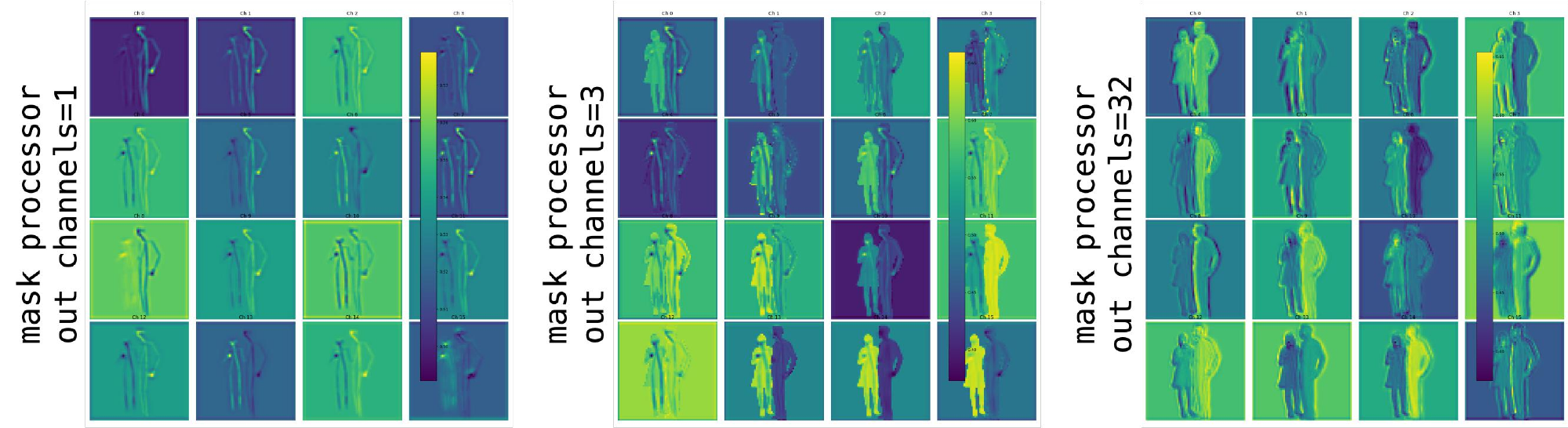}
        \caption{
        The effect of different output channel numbers in the Light mask processor on MaskPoseAdapter output.
        }
        \label{fig:mask_process}
    \end{minipage}
\end{figure}

\subsection{Experiments on residual alpha and mask processor}
Fig.~\ref{fig:residual_alpha} and Fig.~\ref{fig:mask_process} illustrate the influence of various Light mask processors and the parameter $\alpha_{\text{res}}$ on the feature maps generated by MaskPoseAdapter. 
Through extensive experimentation, we determined the optimal number of output channels for the Light mask processor and the value of $\alpha$ that effectively balances the pose and mask features in the output feature maps of MaskPoseAdapter.
In practice, when training on the full dataset, we set $\alpha_{\text{res}}=0.5$ and employ a Light mask processor with 3-channel output.

\section{Data Curation Pipeline}\label{apeendix_datapipe}
Due to the limitations of existing two-person interaction datasets~\cite{swingdance, sun2025beyond}, which fail to simultaneously provide identity diversity, static backgrounds, and fixed camera positions, we propose a novel data processing pipeline that recovers tracked human pose estimations from monocular RGB videos. Our pipeline extracts independent pose sequences, human silhouette masks, and facial masks for distinct individuals. We collected over 170 hours of paired figure skating videos from the internet and curated more than 26 hours of high-quality two-person figure skating segments, providing tracking masks, pose estimations, and facial masks for each individual subject ID. Additionally, we compiled a 1-hour humanoid robot dataset for fine-tuning our model to support controllable video generation tasks involving humanoid robots.

\subsection{Dataset Collection}\label{DatasetCollection}
We collected various single-person motion videos from existing research to enrich identity information, including TikTokDataset~\cite{tiktokdataset}, Champ~\cite{zhu2024champ}, DisPose~\cite{li2025dispose} and HumanVid~\cite{wang2024humanvid}. Additionally, we gathered two-person interaction videos from existing research, including partner dancing, dual talking heads, and laboratory-recorded interactions from Swing Dance~\cite{swingdance}, Harmony4D~\cite{khirodkar2024harmony4d}, HI4D~\cite{yin2023hi4d}, CHI3D~\cite{CHI3D}, and Beyond Talking~\cite{sun2025beyond}. While synthetic data has been used for video generation training in prior work~\cite{yin2024whac, wang2024humanvid}, our method focuses on controllable human interaction video generation in real-world scenarios, so we did not use any synthetic data during training.

\subsection{Human Tracking and Subject Selection}
We first segment raw videos into scenes using TransNetV2~\cite{soucek2024transnet} and detect humans using YOLOv8x~\cite{yolov8_ultralytics}. For each person crop $\mathbf{p}_i^t$, we extract 512-dimensional identity features $\mathbf{f}_i^t$ using pre-trained OSNet~\cite{torchreid,zhou2019osnet,zhou2021osnet}. Our enhanced tracking algorithm combines spatial proximity with ReID similarity to maintain consistent identities across frames. From all tracked identities, we select the two main subjects based on coverage (appearance frequency $\geq 40\%$), consistency, and quality score $Q_i = 0.7 \cdot \text{Coverage}_i + 0.3 \cdot \text{Consistency}_i$.

\subsection{Annotation Generation}
Starting from the bounding boxes ${\mathbf{b}_i^*}$ of key frames, SAMURAI~\cite{yang2024samurai} bidirectionally propagates masks throughout the video sequence. We extract pose information of 133 keypoints using DWPose~\cite{dwpose} and assign each pose to independent subject IDs via an IOU matching approach utilizing the masks generated by SAMURAI. MatAnyone~\cite{yang2025matanyone} produces high-quality alpha mattes from SAMURAI masks, providing data for tasks requiring background replacement~\cite{sun2025beyond}. The complete data processing pipeline is illustrated in Fig.~\ref{fig:data-pipeline}, where our Data Curation Pipeline generates four outputs from RGB video input: independent mask sequences, pose sequences, and facial mask sequences for each individual, as well as alpha mask sequences for all subjects.

\subsection{Data Filtering}
Clips are automatically filtered based on: bbox overlap (max IoU $< 0.1$), size validation ($2\% < $ bbox area $< 80\%$ of frame), exact 2 primary subjects with $\geq 40\%$ coverage, and temporal consistency ($> 90\%$ successful tracking). For PairFS-4K, we additionally perform manual curation to ensure high-quality two-person interactions with clear visibility and balanced representation of skating movements.

\section{PairFS-4K Dataset Preparation Process}\label{PairFS_data_process}
We collected 932 figure skating videos from the internet, including numerous Olympic figure skating compilation videos with multiple shots. Using TransNetV2~\cite{soucek2024transnet}, we developed an automatic segmentation script and employed HumanReID and Yolox for identification and tracking of the main subjects. After manually filtering out segments that did not conform to single-person or pair figure skating criteria, we obtained \textbf{ 4.8K figure skating segments with a total duration of approximately 26 hours, and an average segment length of about 20 seconds}.
We train our model on TikTokDataset~\cite{tiktokdataset}, Champ~\cite{zhu2024champ}, DisPose~\cite{li2025dispose}, HumanVid~\cite{wang2024humanvid}, Swing Dance~\cite{swingdance}, Harmony4D~\cite{khirodkar2024harmony4d}, CHI3D~\cite{CHI3D}, Beyond Talking~\cite{sun2025beyond}, and \textbf{PairFS-4K}, using resolutions of $512\times512$. Due to the limited number of unique identities in HI4D, we exclude it from our training set. 
A detailed summary of all datasets is provided in Table~\ref{tab:datasets-summary}. 
\textbf{PairFS-4K is the first two-person figure skating video dataset with over 7,000 unique identities}.

\section{TogetherVideoBench Benchmark}\label{our_benchmark}
\subsection{Video Generation Benchmark Overview}
There have been many benchmarks for evaluating large generative models~\cite{liu2023fetv, zhang2023zhujiu, sun2024t2v, liu2024evalcrafter, collins2022abo, sun2023journeydb, zhu2023genimage, melistas2024benchmarking, huang2023t2i, iwbench, chang2024survey}. Recently, some video understanding methods have also been used to evaluate the quality of generated videos~\cite{tang2025video, liao2024evaluation, liu2024evalcrafter, ye2024mmad, madan2024foundation}.
Despite this, the field of controllable video generation has lacked a reliable evaluation benchmark. Recent controllable video benchmarks (AIST++~\cite{li2021ai}, TikTok-Eval~\cite{tiktokdataset}) have mainly focused on single-person dance or static portrait animations, overlooking the three key challenges faced by realistic multi-person generation: multi-identity consistency (avoiding identity confusion in long sequences), interaction coherence (ensuring physically reasonable and temporally smooth interactions), and strict conditional fidelity (precisely following pose, mask, or text control inputs).
To systematically evaluate these dimensions, we propose \textbf{TogetherVideoBench}, featuring three orthogonal tracks—\emph{Identity-Consistency}, \emph{Interaction-Coherence}, and \emph{Video Quality}—supported by a unified, automated parsing pipeline that extracts per-person pose, mask, face-crop, and bounding-box representations for fair and reproducible assessment.

\noindent\textbf{Identity-Consistency}: To evaluate the ability of models to maintain consistent appearance and identity for each individual across long video sequences, we adopt standard multi-object tracking metrics, including HOTA~\cite{luiten2021hota}, MOTA~\cite{bernardin2008evaluating}, and IDF1~\cite{ristani2016performance}. These metrics comprehensively assess detection accuracy, association accuracy, and identity preservation, and are computed using the TrackEval toolkit~\cite{luiten2021hota}. This track is crucial for ensuring that generated videos do not suffer from identity switches or appearance confusion, especially in multi-person scenarios.

\noindent\textbf{Interaction-Coherence}: This track focuses on the temporal smoothness and physical plausibility of interactions between multiple humans, as well as the adherence to external control signals. We employ pose adherence (MPJPE-2D)~\cite{cao2017realtime}, object keypoint similarity (OKS)~\cite{lin2014coco}, and the following metrics: pose structure similarity (PoseSSIM), motion smoothness (SmoothRMS), temporal dynamics error (TimeDynRMSE), and Fréchet Video Motion Distance (FVMD)~\cite{liu2024fr} to comprehensively evaluate the quality of human motion and interaction.

\begin{figure}[htbp]
    \centering
    \includegraphics[width=1\linewidth]{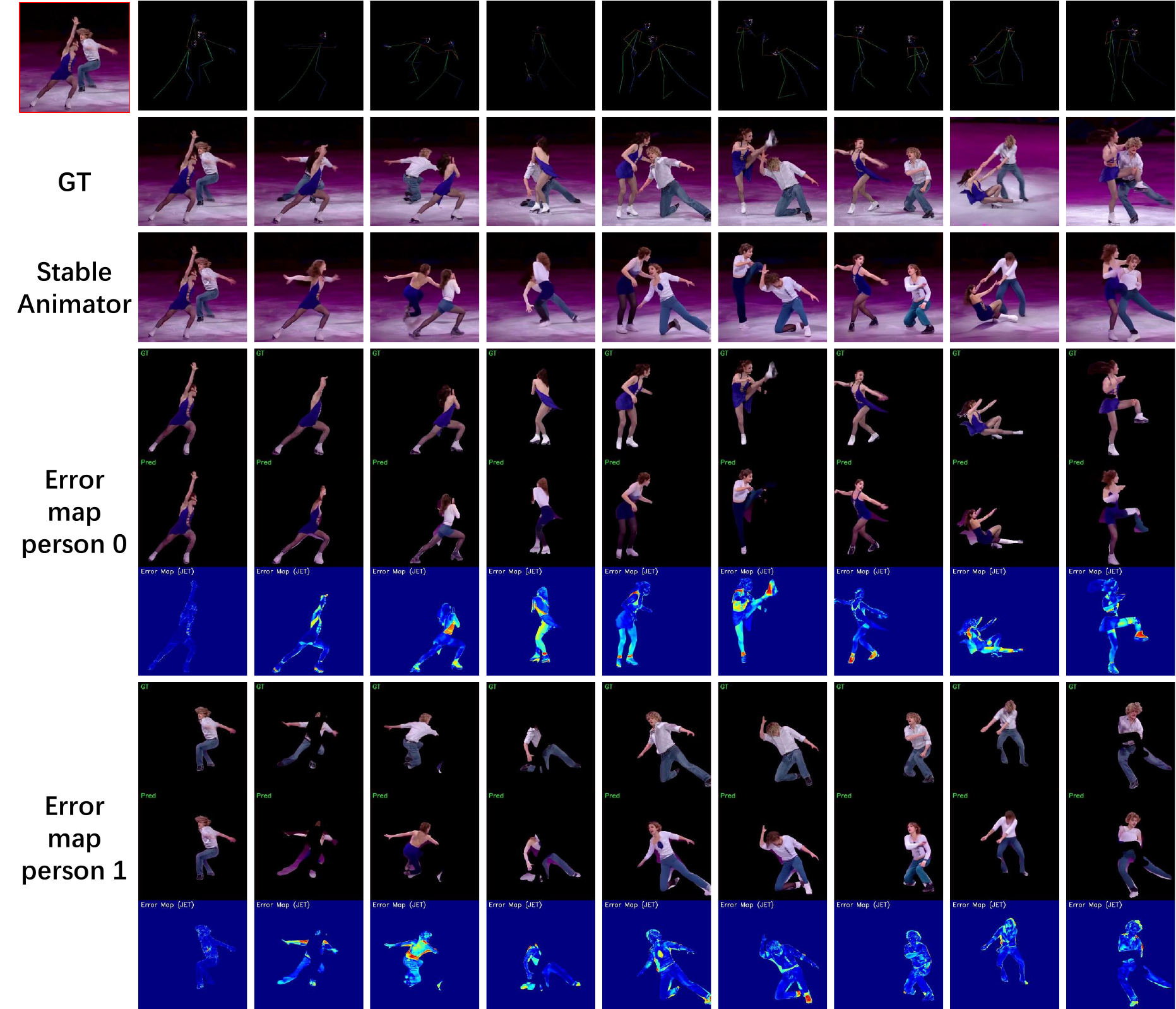}
    \caption{
    We use individual human masks for each person to conduct quantitative evaluation. The error map shown in the figure is the L1 Loss error map, which calculates the pixel-level absolute difference between the GT and predicted images.
    }
    \label{fig:error_map}
\end{figure}

\noindent\textbf{Video Quality}: To assess the overall visual fidelity and semantic consistency of generated videos, we use a suite of widely adopted metrics, including SSIM~\cite{1284395}, FVD~\cite{unterthiner2019fvd}, FID~\cite{NIPS2017_8a1d6947},  CLIP~\cite{hessel2021clipscore}, and the following metrics: LPIPS, L1, PSNR, DISTS~\cite{ding2020image}, ST-SSIM~\cite{moorthy2010efficient}, GMSD-T~\cite{yan2015video}. These metrics collectively measure both the perceptual quality and the alignment of generated content with the intended conditions.
We calculate the metrics for both the overall frame and the human mask region of each frame separately, as shown in Fig.~\ref{fig:error_map}. Since the backgrounds of some evaluation data exhibit slight jittering, we believe that the quantitative evaluation of the human mask region is more indicative of human ID consistency and video quality in the generated videos than the quantitative evaluation of the full frame.

All tracks share a unified Data Curation Pipeline that automatically extracts per-person pose, mask, face-crop, and bounding box for both ground truth and generated videos, ensuring reproducibility and fair comparison. For each video, we compute the relevant metrics for every individual and report the average across all videos in each group. 

\subsection{Evaluation Dataset}\label{subsec:Evaluation_Dataset}
While laboratory-recorded datasets such as Harmony4D~\cite{khirodkar2024harmony4d}, HI4D~\cite{yin2023hi4d}, and CHI3D~\cite{CHI3D} provide precise annotations, their videos are typically limited to 3–12 seconds, feature single scenes, and involve minimal position exchanges between subjects. As a result, they are insufficient for evaluating long-duration, multi-position, and realistic human interactions. 
To address this gap, we have manually curated and edited 100 high-quality two-person interaction videos from public competitions, films, documentaries, and social media, forming the core evaluation set of \texttt{TogetherVideoBench}. 
These videos encompass a wide range of real-world interaction patterns, including exchange-intensive swing and Lindy-Hop routines, Latin ballroom duets, pair figure skating, boxing, wrestling and combat sequences, partner acrobatics and acro-yoga throws, everyday social gestures (such as handshakes and hugs), and two-person conversations. 
Each clip features exactly two performers, with nearly static cameras and backgrounds. Frequent occlusions, position exchanges, and physical contact between subjects introduce long-range motion, viewpoint changes, and identity-switching challenges—factors, making it a suitable testbed. 

\subsection{Metrics}
\label{Metrics}
Below are the evaluation metrics and computation procedures used in the three tracks of TogetherVideoBench.
To ensure reproducibility, both ground-truth and generated videos are first processed by our Data Curation Pipeline (Sec.~\ref{apeendix_datapipe}), which yields for each subject:
\begin{itemize}[leftmargin=1.4em,itemsep=1pt]
  \item \textbf{Pose sequences:} 133 keypoints per frame via DWPose~\cite{dwpose}.  
  \item \textbf{Human masks:} per-frame human masks via SAMURAI~\cite{yang2024samurai}.
  \item \textbf{Bounding boxes:} tight boxes around each human mask (for MOT eval~\cite{luiten2021hota}).  
\end{itemize}

\paragraph{Track 1 – Identity-Consistency}
\begin{itemize}[leftmargin=1.4em,itemsep=1pt]
    \item \textbf{IDF1\,↑}:\\
    After frame–level association with the Hungarian algorithm, let $\mathrm{IDTP}$, $\mathrm{IDFP}$ and $\mathrm{IDFN}$ be identity–true positives, false positives and false negatives.
    \begin{equation}
    \mathrm{IDF1} = \frac{2\,|\mathrm{IDTP}|}{2\,|\mathrm{IDTP}| + |\mathrm{IDFP}| + |\mathrm{IDFN}|}.
    \end{equation}
    It is the harmonic mean of identity precision and recall and therefore measures how often the \emph{correct ID label} is maintained.

    \item \textbf{IDP / IDR\,↑}:\\
    Precision and recall components of IDF1.
    \begin{equation}
    \mathrm{IDP} = \frac{|\mathrm{IDTP}|}{|\mathrm{IDTP}| + |\mathrm{IDFP}|}, 
    \quad
    \mathrm{IDR} = \frac{|\mathrm{IDTP}|}{|\mathrm{IDTP}| + |\mathrm{IDFN}|}.
    \end{equation}

    \item \textbf{HOTA\,↑}:\\
    Higher-Order Tracking Accuracy~\cite{luiten2021hota} decomposes into $\mathrm{DetA}$ (detection accuracy), $\mathrm{AssA}$ (association accuracy) and $\mathrm{LocA}$ (localisation accuracy):
    \begin{equation}
    \mathrm{HOTA} = \sqrt{\mathrm{DetA} \times \mathrm{AssA}},
    \quad
    \mathrm{LocA} = 1 \;-\; \frac{1}{|\mathrm{TP}|}\sum_{b \in \mathrm{TP}}\bigl(1 - \mathrm{IoU}(b)\bigr).
    \end{equation}

    \item \textbf{MOTA / MOTP\,↑}:\\
    CLEAR-MOT summary:
    \begin{equation}
    \mathrm{MOTA} = 1 \;-\; \frac{\mathrm{FP} + \mathrm{FN} + \mathrm{IDSW}}{\text{GT dets}},
    \quad
    \mathrm{MOTP} = 1 \;-\; \frac{\sum_{\mathrm{TP}}\bigl(1 - \mathrm{IoU}\bigr)}{|\mathrm{TP}|}.
    \end{equation}

    \item \textbf{IDSW\,↓}, \textbf{FP\,↓}, \textbf{FN\,↓}:\\
    Absolute counts of identity switches, false positives and false negatives.
\end{itemize}

\paragraph{Track 2 – Interaction-Coherence}
All keypoints are first temporally aligned and isotropically scale–shift aligned via a similarity transform.

\begin{itemize}[leftmargin=1.4em,itemsep=1pt]
  \item \textbf{MPJPE-2D\,↓}:\\
  Let $\hat{\mathbf{x}}_{tpj}$ and $\mathbf{x}_{tpj}$ be the predicted and ground‐truth pixel coordinates of joint $j$ of person $p$ at frame $t$, after SIM3 alignment; $T,P,J$ denote total frames, persons, and joints. Then
  \begin{equation}
  \mathrm{MPJPE\text{-}2D}
  = \frac{1}{T\,P\,J}
  \sum_{t=1}^T \sum_{p=1}^P \sum_{j=1}^J
  \bigl\lVert \hat{\mathbf{x}}_{tpj} - \mathbf{x}_{tpj} \bigr\rVert_2.
  \end{equation}

  \item \textbf{OKS\,↑}:\\
  For each frame $t$, flatten over $P\times J$ valid keypoints. Let $d_k$ be the Euclidean error of the $k$th keypoint, $\sigma_k$ its COCO standard deviation, and $\mathcal{A}$ the estimated person area. Then
  \begin{equation}
  \mathrm{OKS}_t
  = \frac{1}{K}
  \sum_{k=1}^K
    \exp\!\Bigl(-\frac{d_k^2}{2\,\sigma_k^2(\mathcal{A}+10^{-6})}\Bigr),
  \qquad
  \mathrm{OKS} = \frac{1}{T}\sum_{t=1}^T \mathrm{OKS}_t.
  \end{equation}

  \item \textbf{Pose-Heat SSIM\,↑}:\\
  Rasterise the set of keypoints at each frame into a Gaussian heatmap $H(\cdot)$ of size $H\times W$ with $\sigma=4\,$px, then
  \begin{equation}
  \mathrm{PoseHeatSSIM}
  = \frac{1}{T}\sum_{t=1}^T
    \mathrm{SSIM}\bigl(H(\hat{\mathbf{X}}_t),\,H(\mathbf{X}_t)\bigr),
  \end{equation}
  where $\hat{\mathbf{X}}_t,\mathbf{X}_t\in\mathbb{R}^{P\times J\times 2}$ are the keypoint arrays.

  \item \textbf{SmoothRMS\,↓}:\\
  Compute the third‐order temporal derivative (jerk) of each trajectory, scaled by frame rate $f$:
  \begin{equation}
    \dddot{\mathbf{x}}_{tpj}
    = \frac{d^3}{dt^3}\mathbf{x}_{tpj}\,\times f^3.
  \end{equation}
  Then
  \begin{equation}
  \mathrm{SmoothRMS}
  = \sqrt{
      \frac{1}{T\,P\,J}
      \sum_{t=1}^T \sum_{p=1}^P \sum_{j=1}^J
      \bigl\lVert \dddot{\mathbf{x}}_{tpj}\bigr\rVert_2^2
    }.
  \end{equation}

  \item \textbf{Time-Dyn RMSE\,↓}:\\
  With the second‐order derivative (acceleration)
  \begin{equation}
    \ddot{\mathbf{x}}_{tpj}
    = \frac{d^2}{dt^2}\mathbf{x}_{tpj}\,\times f^2,
  \end{equation}
  define
  \begin{equation}
  \mathrm{TimeDynRMSE}
  = \sqrt{
      \frac{1}{T\,P\,J}
      \sum_{t=1}^T \sum_{p=1}^P \sum_{j=1}^J
      \bigl\lVert \ddot{\mathbf{x}}_{tpj}\bigr\rVert_2^2
    }.
  \end{equation}

  \item \textbf{FVMD\,↓}:\\
  Model the velocity vectors of all keypoints as 2D Gaussians
  $\mathcal{N}(\mu_p,\Sigma_p)$ for prediction and $\mathcal{N}(\mu_g,\Sigma_g)$ for ground truth,
  where
  $\mu = \mathbb{E}[\dot{\mathbf{x}}]$ and
  $\Sigma=\operatorname{Cov}[\dot{\mathbf{x}}]$. Then
  \begin{equation}
  \mathrm{FVMD}
  = \bigl\lVert \mu_p - \mu_g\bigr\rVert_2^2
   \;+\;\operatorname{Tr}\bigl(\Sigma_p + \Sigma_g - 2\,(\Sigma_p\,\Sigma_g)^{\tfrac12}\bigr).
  \end{equation}
\end{itemize}

\paragraph{Track 3 – Video Quality}
\begin{itemize}[leftmargin=1.4em,itemsep=1pt]
    \item \textbf{L1\,↓}:\\
    Let $I_{t}(x,y,c)$ and $\hat I_{t}(x,y,c)$ be the ground-truth and predicted RGB pixel values at frame $t$, spatial location $(x,y)$ and channel $c$, over $T$ frames of size $H\times W$ and $C=3$ channels. Then
    \begin{equation}
      \mathrm{L1}
      = \frac{1}{T\,H\,W\,C}
      \sum_{t=1}^{T}\sum_{x=1}^{W}\sum_{y=1}^{H}\sum_{c=1}^{C}
      \bigl\lvert I_{t}(x,y,c) - \hat I_{t}(x,y,c)\bigr\rvert.
    \end{equation}

    \item \textbf{PSNR\,↑}:\\
    Compute the per-frame mean squared error
    \begin{equation}
      \mathrm{MSE}
      = \frac{1}{H\,W\,C}
        \sum_{x=1}^{W}\sum_{y=1}^{H}\sum_{c=1}^{C}
        \bigl(I_{t}(x,y,c)-\hat I_{t}(x,y,c)\bigr)^2,
    \end{equation}
    then
    \begin{equation}
      \mathrm{PSNR}
      = 20\log_{10}\!\Bigl(\frac{255}{\sqrt{\mathrm{MSE}}}\Bigr).
    \end{equation}

    \item \textbf{SSIM\,↑}:\\
    For each frame $t$ and each channel $c$, compute
    \begin{equation}
      \mathrm{SSIM}_t^c
      = \mathrm{SSIM}\bigl(I_{t}(\cdot,\cdot,c),\,\hat I_{t}(\cdot,\cdot,c)\bigr),
    \end{equation}
    then average:
    \begin{equation}
      \mathrm{SSIM}
      = \frac{1}{T\,C}
      \sum_{t=1}^T \sum_{c=1}^C \mathrm{SSIM}_t^c.
    \end{equation}

    \item \textbf{LPIPS\,↓}:\\
    On a $256\times256$ crop, let $\phi_\ell(\cdot)$ be the $\ell$-th layer feature map and $w_\ell$ learned weights. Then
    \begin{equation}
      \mathrm{LPIPS}
      = \frac{1}{L}
      \sum_{\ell=1}^{L}
      \frac{1}{H_\ell W_\ell}
      \bigl\lVert\,w_\ell \,\odot\,
      \bigl(\phi_\ell(I)-\phi_\ell(\hat I)\bigr)\bigr\rVert_1.
    \end{equation}

    \item \textbf{DISTS\,↓}:\\
    Let $f_\ell(\cdot)$ be VGG16 feature maps, $\tilde f_\ell$ their normalized versions, and $G(\cdot)$ the Gram matrix. Define
    \begin{equation}
      \mathrm{structure}_\ell
      = \frac{\langle\tilde f_\ell(I),\,\tilde f_\ell(\hat I)\rangle}
             {\|\tilde f_\ell(I)\|\,\|\tilde f_\ell(\hat I)\|},
      \quad
      \mathrm{texture}_\ell
      = \mathrm{MSE}\bigl(G(f_\ell(I)),\,G(f_\ell(\hat I))\bigr).
    \end{equation}
    Then
    \begin{equation}
      \mathrm{DISTS}
      = \frac{1}{L}
      \sum_{\ell=1}^{L}
      \Bigl(
        0.5\,\mathrm{structure}_\ell
        +0.5\,\bigl(1-\mathrm{texture}_\ell\bigr)
      \Bigr).
    \end{equation}

    \item \textbf{CLIPScore\,↑}:\\
    We encode each frame from the ground-truth and generated videos into CLIP image embeddings $v_t, \hat{v}_t \in \mathbb{R}^d$, normalize them to unit vectors, and compute the frame-wise cosine similarity:
    \begin{equation}
    s_t = \frac{v_t^\top \hat{v}_t}{\|v_t\| \cdot \|\hat{v}_t\|}.
    \end{equation}
    The final CLIPScore is obtained by averaging over all $T$ frames:
    \begin{equation}
    \mathrm{CLIPScore} = \frac{1}{T} \sum_{t=1}^{T} s_t.
    \end{equation}

    \item \textbf{ST-SSIM\,↑}:\\
    With window length $w=3$, define for each spatio-temporal block
    \begin{equation}
      \mathrm{SSIM}_{3\mathrm D}
      = \mathrm{SSIM}\bigl(I_{t:t+w-1},\,\hat I_{t:t+w-1}\bigr),
    \end{equation}
    then
    \begin{equation}
      \mathrm{ST\text{-}SSIM}
      = \frac{1}{T-w+1}
      \sum_{t=1}^{T-w+1}
      \mathrm{SSIM}_{3\mathrm D}.
    \end{equation}

    \item \textbf{GMSD-Temporal\,↓}:\\
    For each $t=2,\dots,T$, let
    \begin{equation}
      g_t(x,y)
      = \bigl\lVert\nabla I_t(x,y)\bigr\rVert_2,\quad
      \hat g_t(x,y)
      = \bigl\lVert\nabla \hat I_t(x,y)\bigr\rVert_2,
    \end{equation}
    and
    \begin{equation}
      \mathrm{GMS}_t(x,y)
      = \frac{2\,g_t\,\hat g_t+\varepsilon}{g_t^2+\hat g_t^2+\varepsilon}.
    \end{equation}
    Then
    \begin{equation}
      \mathrm{GMSD\text{-}Temporal}
      = \sqrt{
        \frac{1}{(T-1)\,H\,W}
        \sum_{t=2}^T
        \mathrm{Var}_{x,y}\!\bigl(\mathrm{GMS}_t(x,y)\bigr)
      }.
    \end{equation}

    \item \textbf{FVD\,↓}:\\
    Extract I3D features for each non-overlapping 16-frame clip, compute means $\mu_r,\mu_f$ and covariances $\Sigma_r,\Sigma_f$, then
    \begin{equation}
      \mathrm{FVD}
      = \bigl\lVert \mu_r - \mu_f \bigr\rVert_2^2
      + \mathrm{Tr}\bigl(\Sigma_r + \Sigma_f - 2(\Sigma_r\Sigma_f)^{1/2}\bigr).
    \end{equation}

    \item \textbf{FID\,↓}:\\
    On all frames, extract Inception-V3 features, form $(\mu_r,\Sigma_r)$ and $(\mu_f,\Sigma_f)$, and use
    \begin{equation}
      \mathrm{FID}
      = \bigl\lVert \mu_r - \mu_f \bigr\rVert_2^2
      + \mathrm{Tr}\bigl(\Sigma_r + \Sigma_f - 2(\Sigma_r\Sigma_f)^{1/2}\bigr).
    \end{equation}

    \item \textbf{CLIP-FID\,↓}:\\
    Identical to FID but using CLIP embeddings instead of Inception features:
    \begin{equation}
      \mathrm{CLIP\!-\!FID}
      = \bigl\lVert \mu_r - \mu_f \bigr\rVert_2^2
      + \mathrm{Tr}\bigl(\Sigma_r + \Sigma_f - 2(\Sigma_r\Sigma_f)^{1/2}\bigr).
    \end{equation}
\end{itemize}

\medskip
\noindent
All Track-3 metrics are reported both on the \emph{full frame} and on each human mask (per-person); the final masked score is the arithmetic mean over the two performers.

\section{More Results}\label{MoreResults}
\begin{figure}[htbp]
    \centering
    \includegraphics[width=1\linewidth]{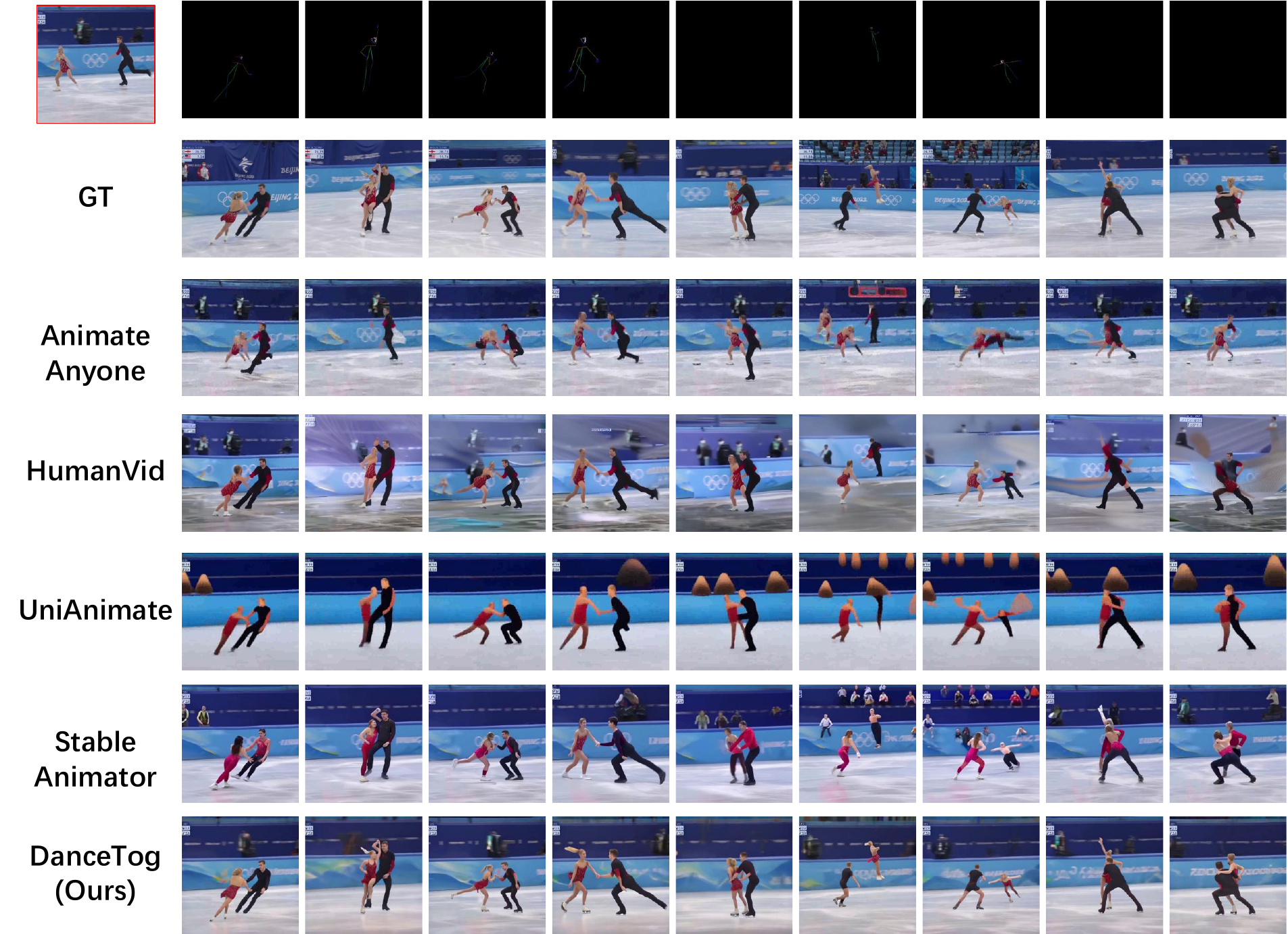}
    \caption{Additional animation results (1/4). The image with red borders is the reference images.}
    \label{fig:animation_result}
\end{figure}

\begin{figure}[htbp]
    \centering
    \includegraphics[width=1\linewidth]{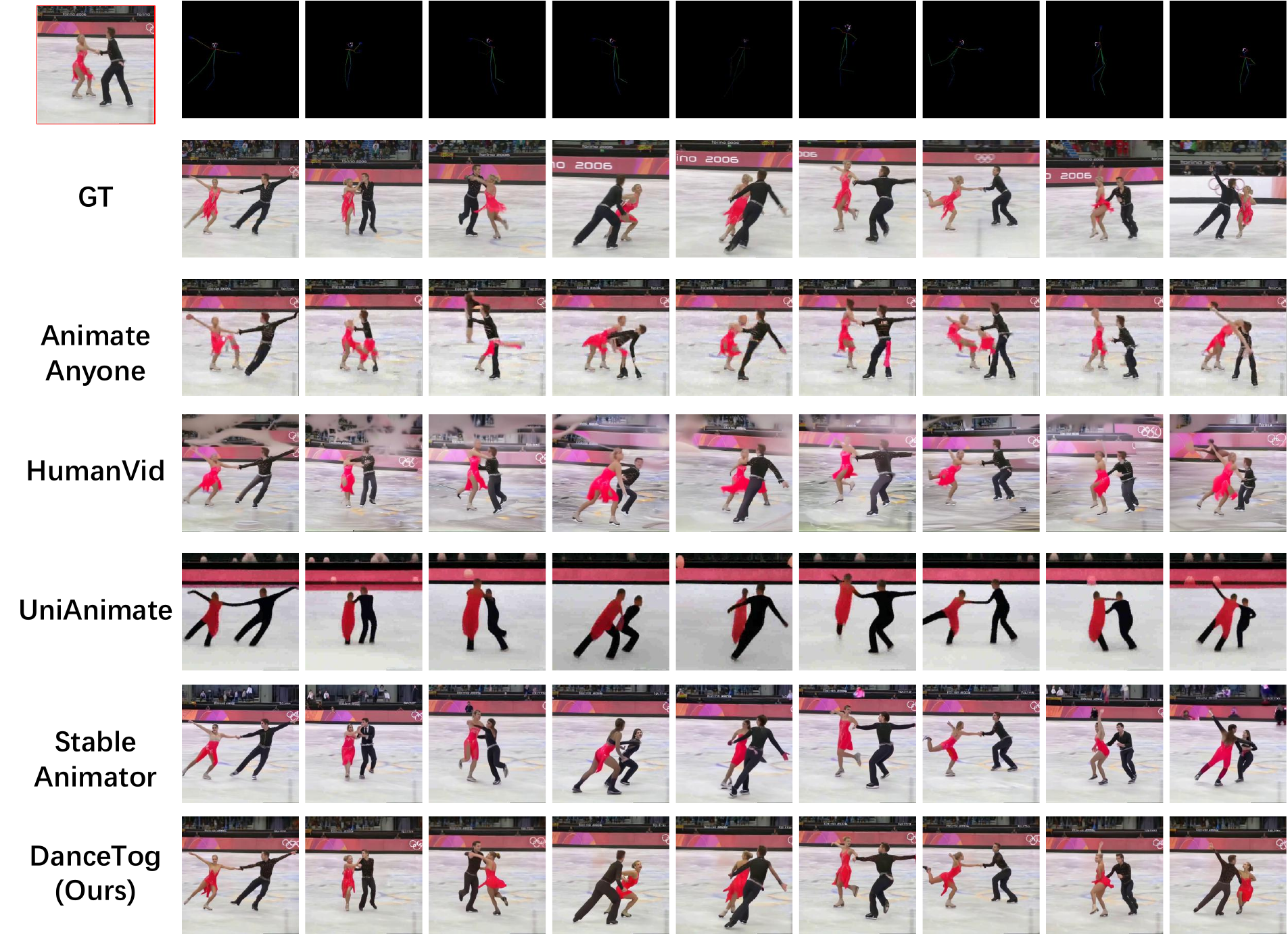}
    \caption{Additional animation results (2/4). The image with red borders is the reference images.}
    \label{fig:animation_result2}
\end{figure}

\begin{figure}[htbp]
    \centering
    \includegraphics[width=1\linewidth]{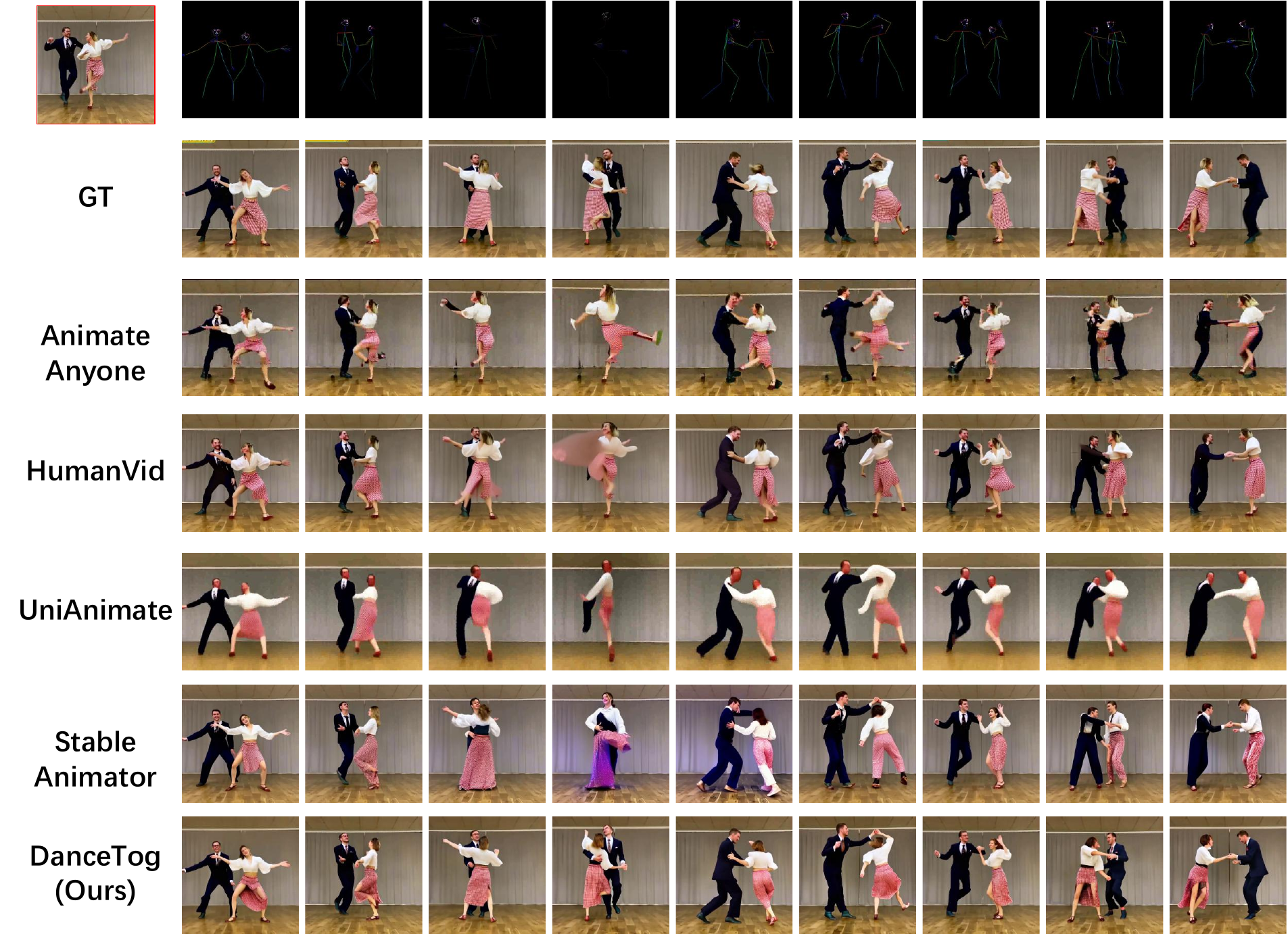}
    \caption{Additional animation results (3/4). The image with red borders is the reference images.}
    \label{fig:animation_result3}
\end{figure}

\begin{figure}[htbp]
    \centering
    \includegraphics[width=1\linewidth]{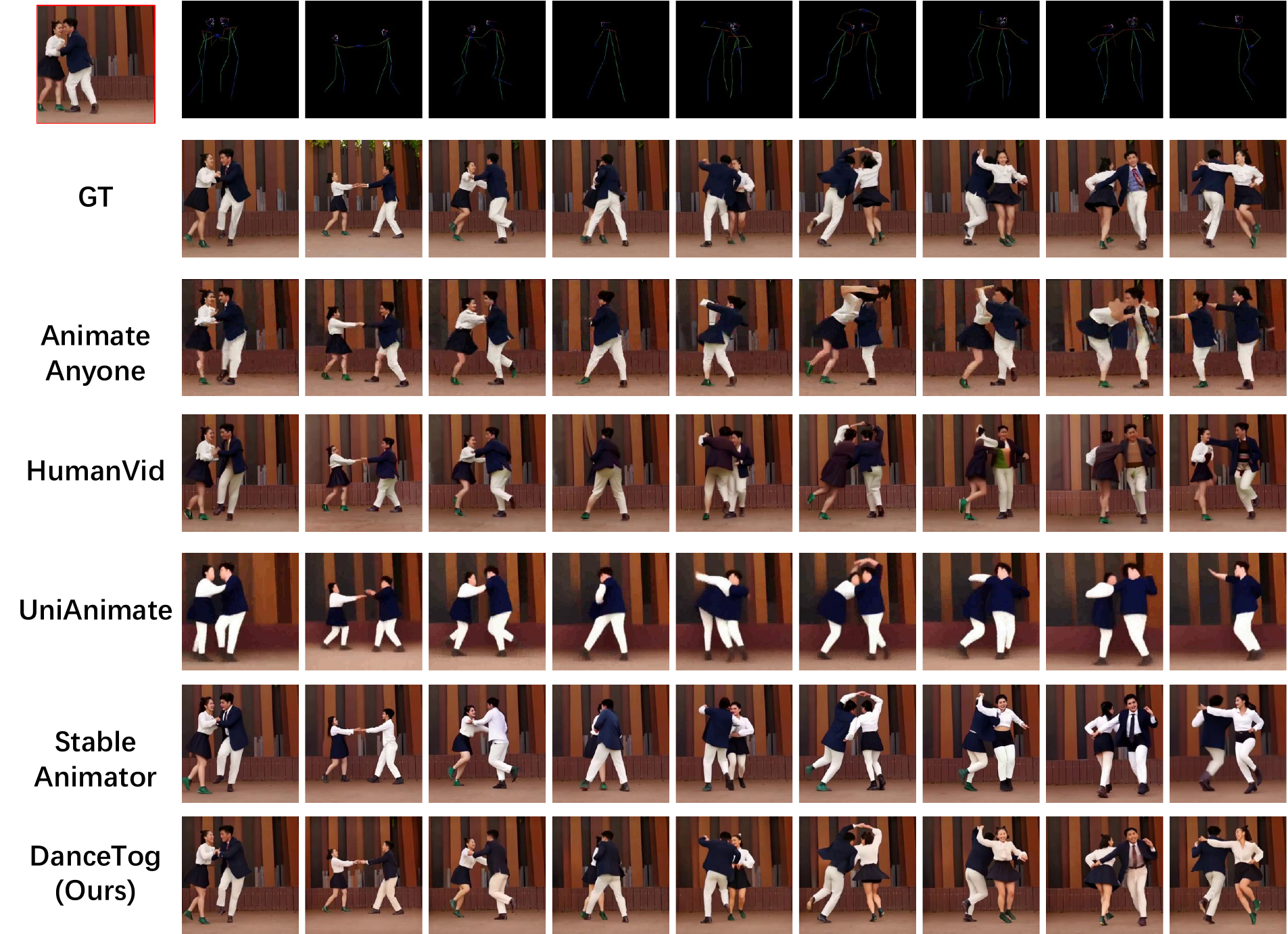}
    \caption{Additional animation results (4/4). The image with red borders is the reference images.}
    \label{fig:animation_result4}
\end{figure}

Fig.~\ref{fig:animation_result}, Fig.~\ref{fig:animation_result2}, Fig.~\ref{fig:animation_result3}, and Fig.~\ref{fig:animation_result4} present qualitative comparisons across consecutive frames for different cases. The top row in each figure shows the input reference image and the corresponding pose sequence. The pose sequence is estimated from a ground truth video, and the first frame is used as the reference image input for each baseline.
Our proposed DanceTog method consistently outperforms all baselines in generating video frames with rich interaction details. Notably, it preserves individual identity features even when the two subjects exchange positions.
For qualitative video comparisons, please refer to the supplementary webpage.

\begin{figure}[htbp]
\centering
\includegraphics[width=1\linewidth]{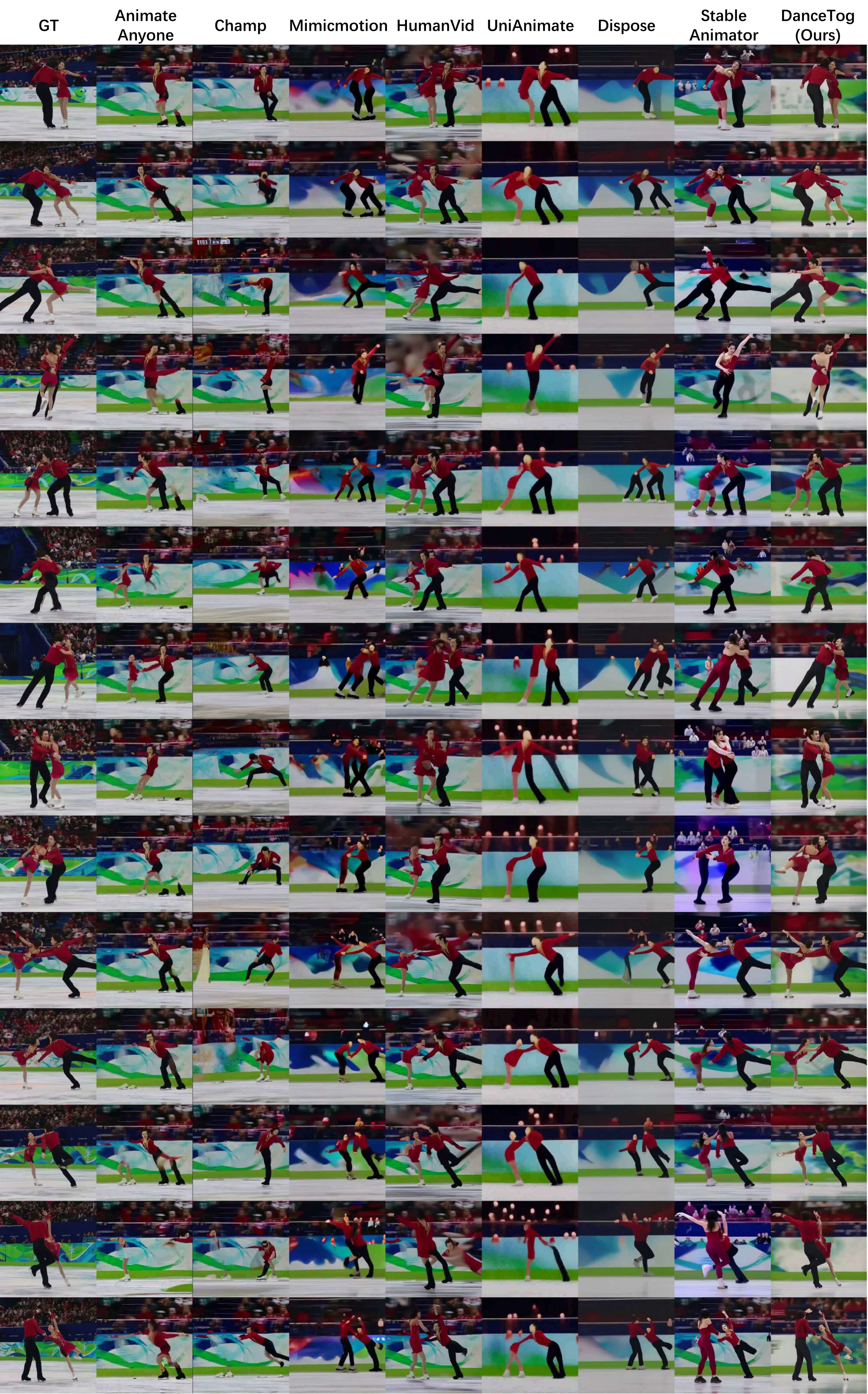}
\caption{Additional animation results (4/18). }
\label{fig:supp-case04}
\end{figure}

\begin{figure}[htbp]
\centering
\includegraphics[width=1\linewidth]{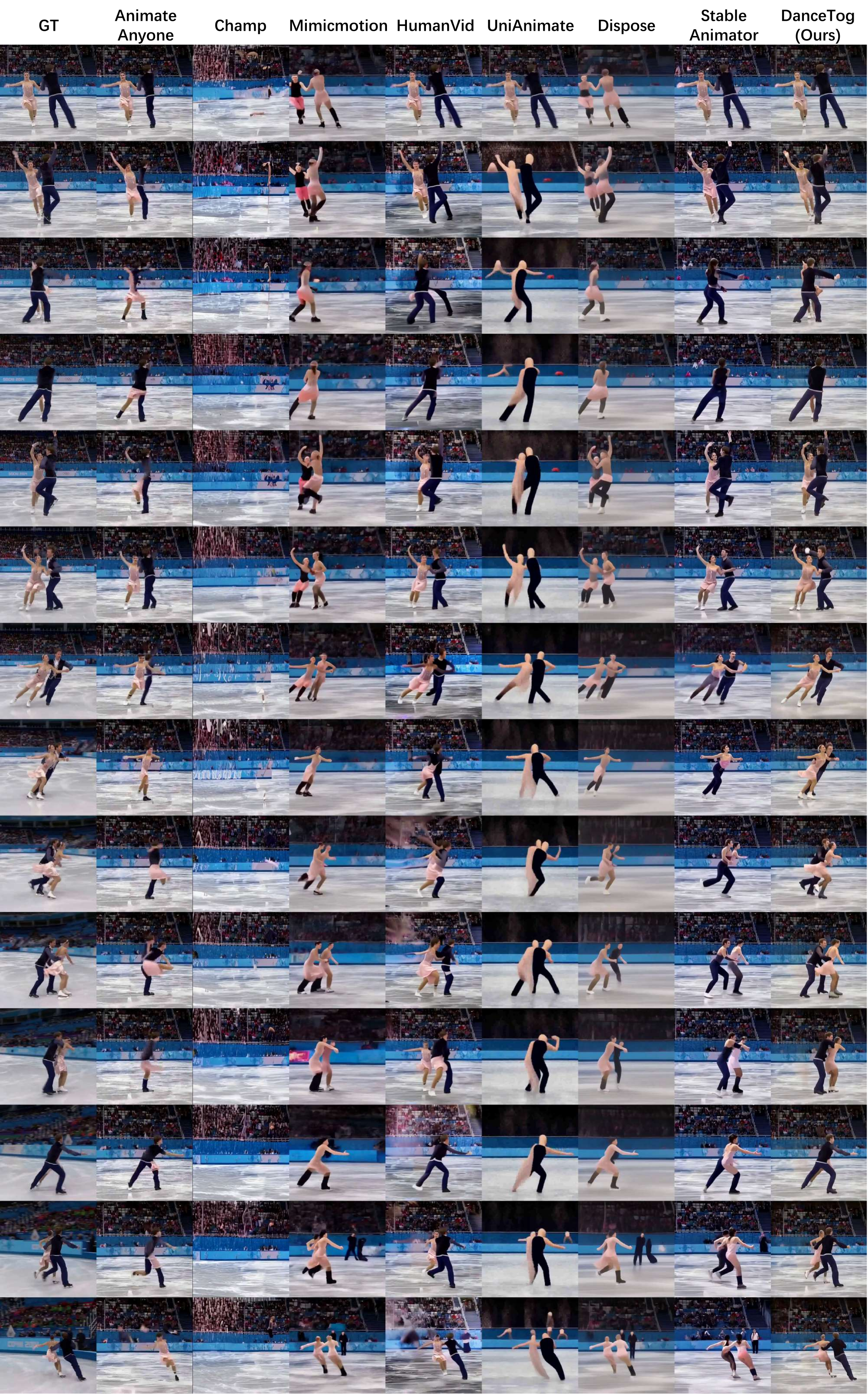}
\caption{Additional animation results (5/18). }
\label{fig:supp-case05}
\end{figure}

\begin{figure}[htbp]
\centering
\includegraphics[width=1\linewidth]{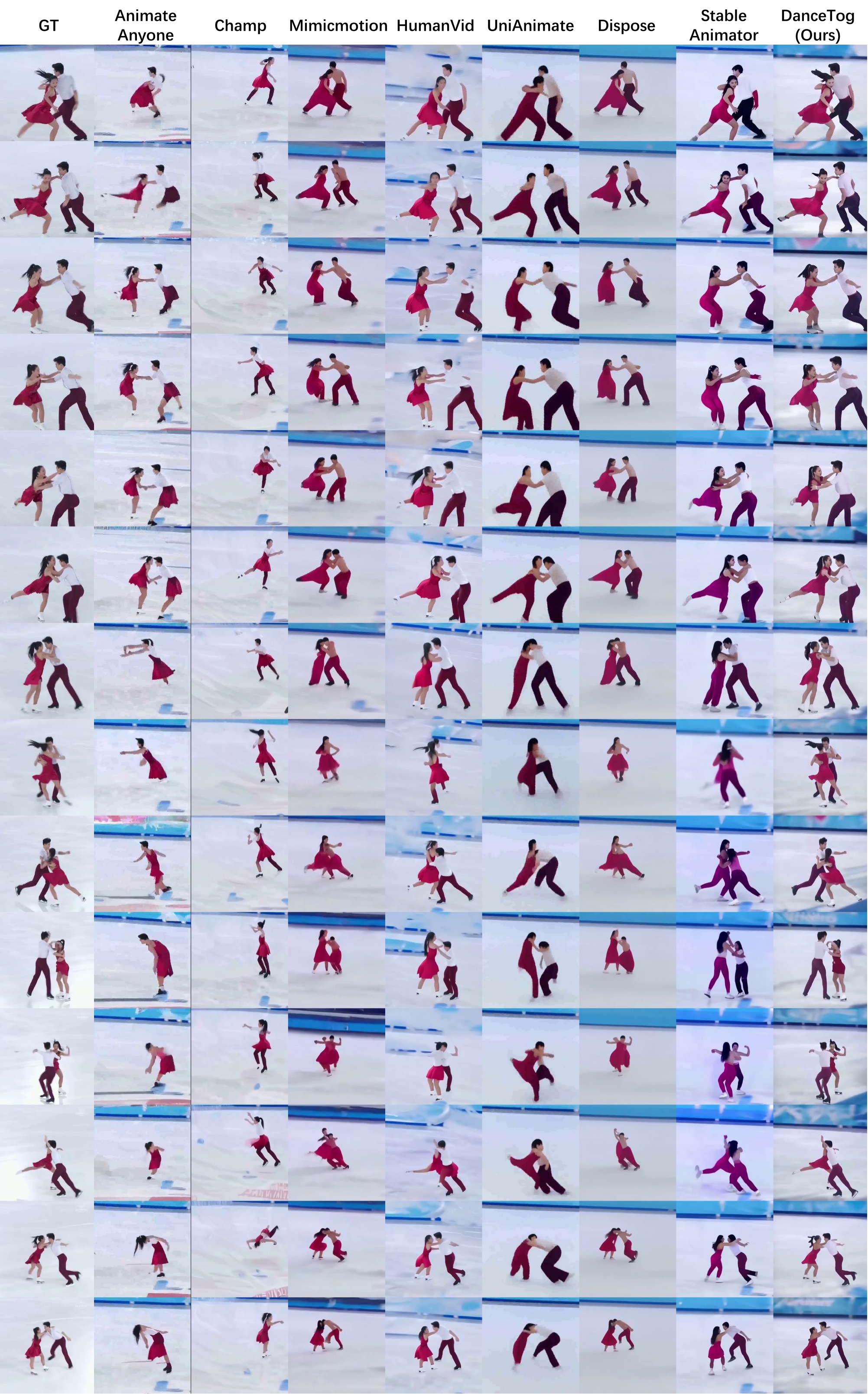}
\caption{Additional animation results (8/18). }
\label{fig:supp-case08}
\end{figure}

\begin{figure}[htbp]
\centering
\includegraphics[width=1\linewidth]{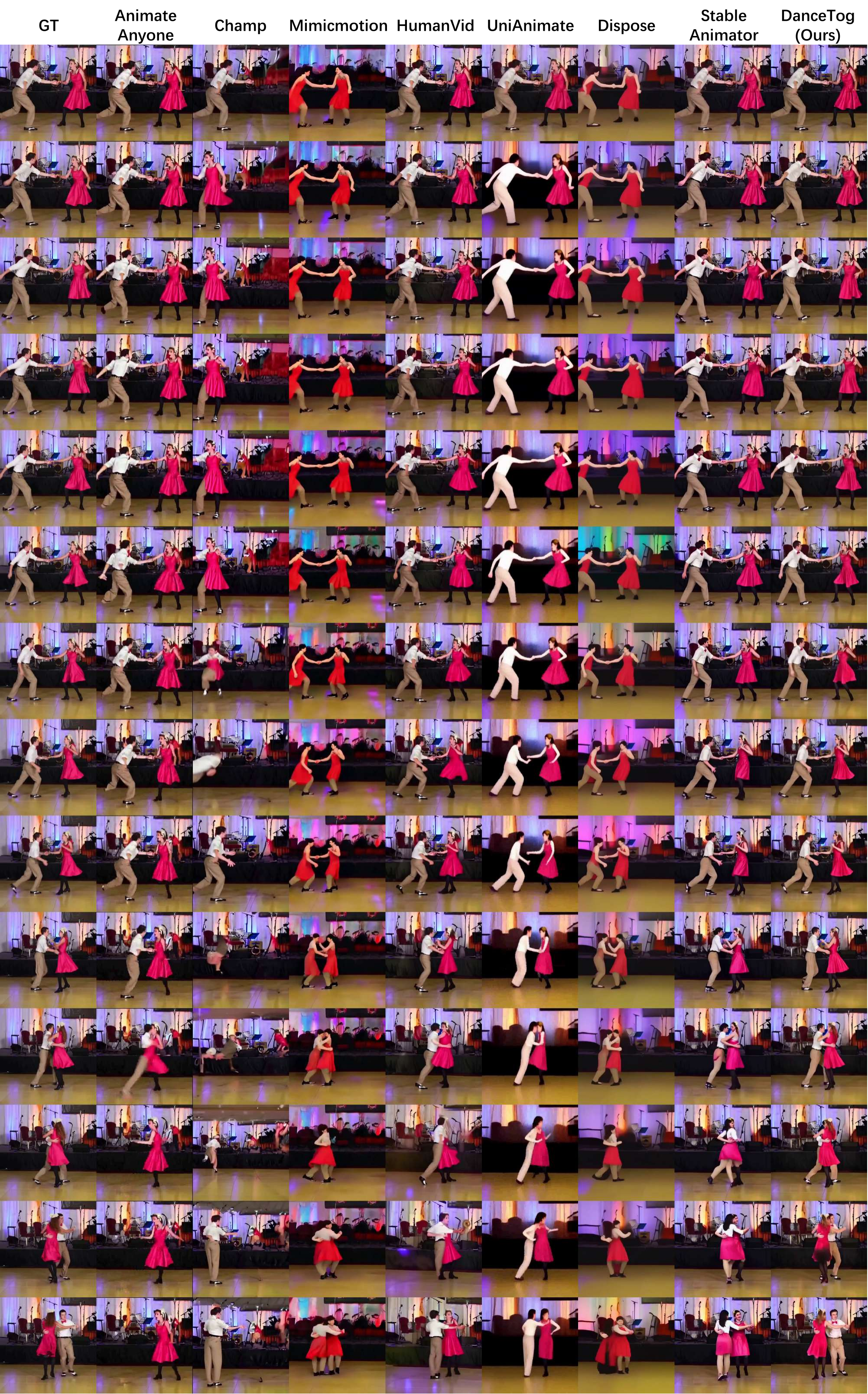}
\caption{Additional animation results (10/18). }
\label{fig:supp-case10}
\end{figure}

\begin{figure}[htbp]
\centering
\includegraphics[width=1\linewidth]{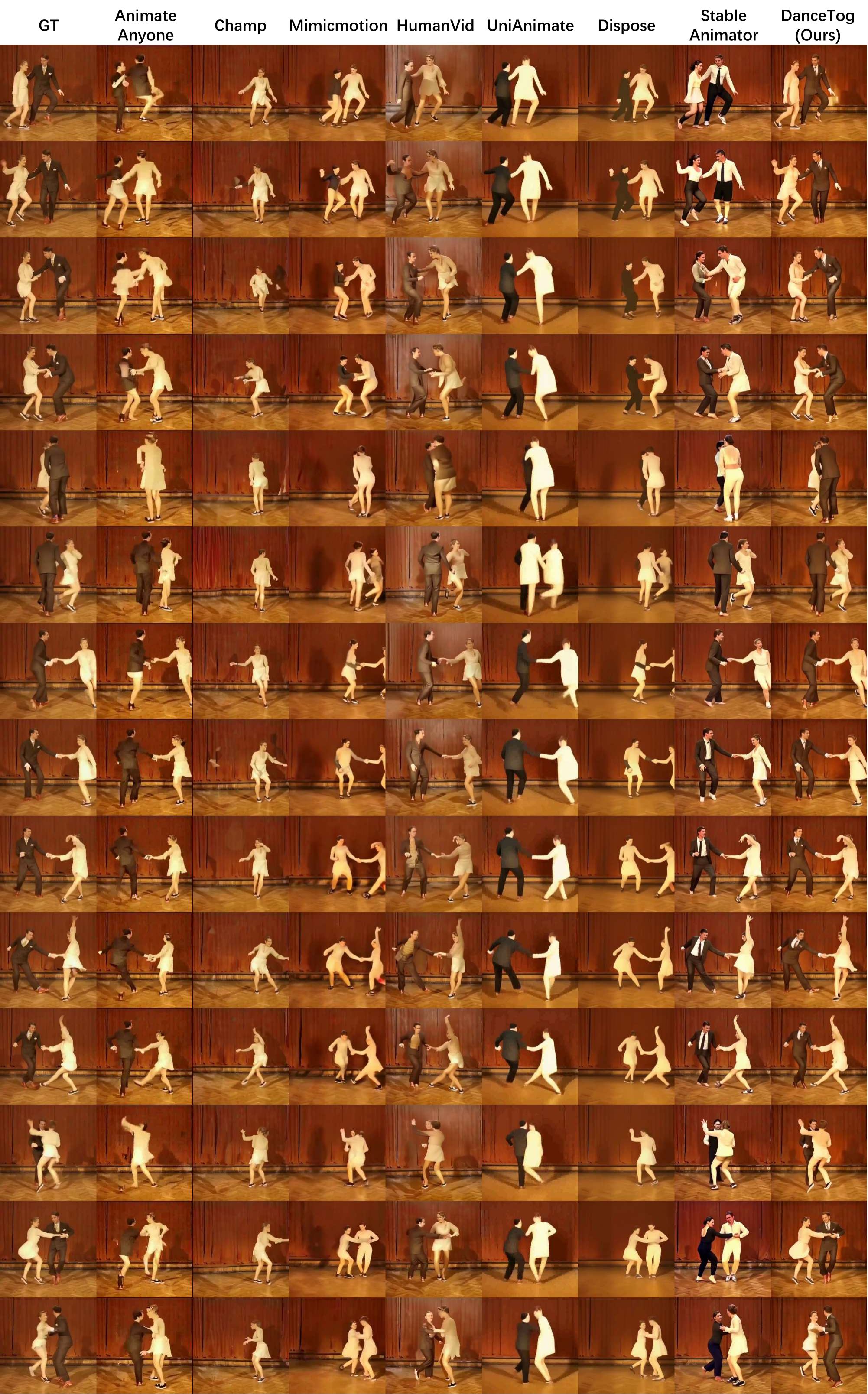}
\caption{Additional animation results (11/18). }
\label{fig:supp-case11}
\end{figure}

\begin{figure}[htbp]
\centering
\includegraphics[width=1\linewidth]{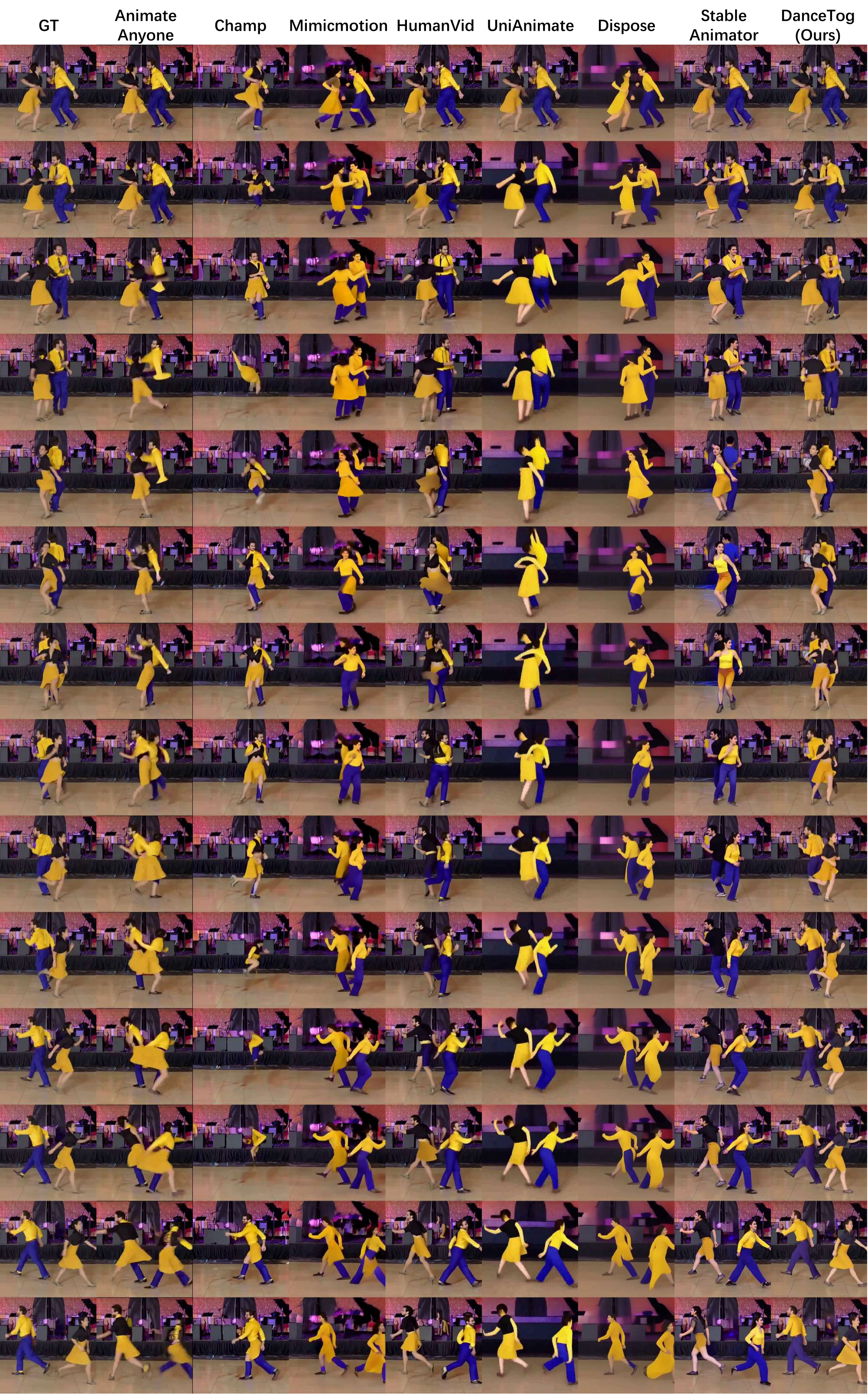}
\caption{Additional animation results (12/18). }
\label{fig:supp-case12}
\end{figure}

Figs.~\ref{fig:supp-case04}--\ref{fig:supp-case12} show qualitative comparisons of all baselines. We extracted consecutive frames where position swapping occurs. The leftmost column is the GT video. We used the first frame of the GT video as the reference image (not the first frame shown in the figures), and the dwpose results estimated from the GT video as the pose condition input for each baseline (corresponding to the GT images in the first column). Due to file size limitations, the images below are compressed. Please refer to the webpage in the supplementary materials for the original videos.

\section{Applications: Human–Robot Interaction Video Generation}\label{app_HRI}
\begin{figure}[htbp]
    \centering
    \includegraphics[width=1\linewidth]{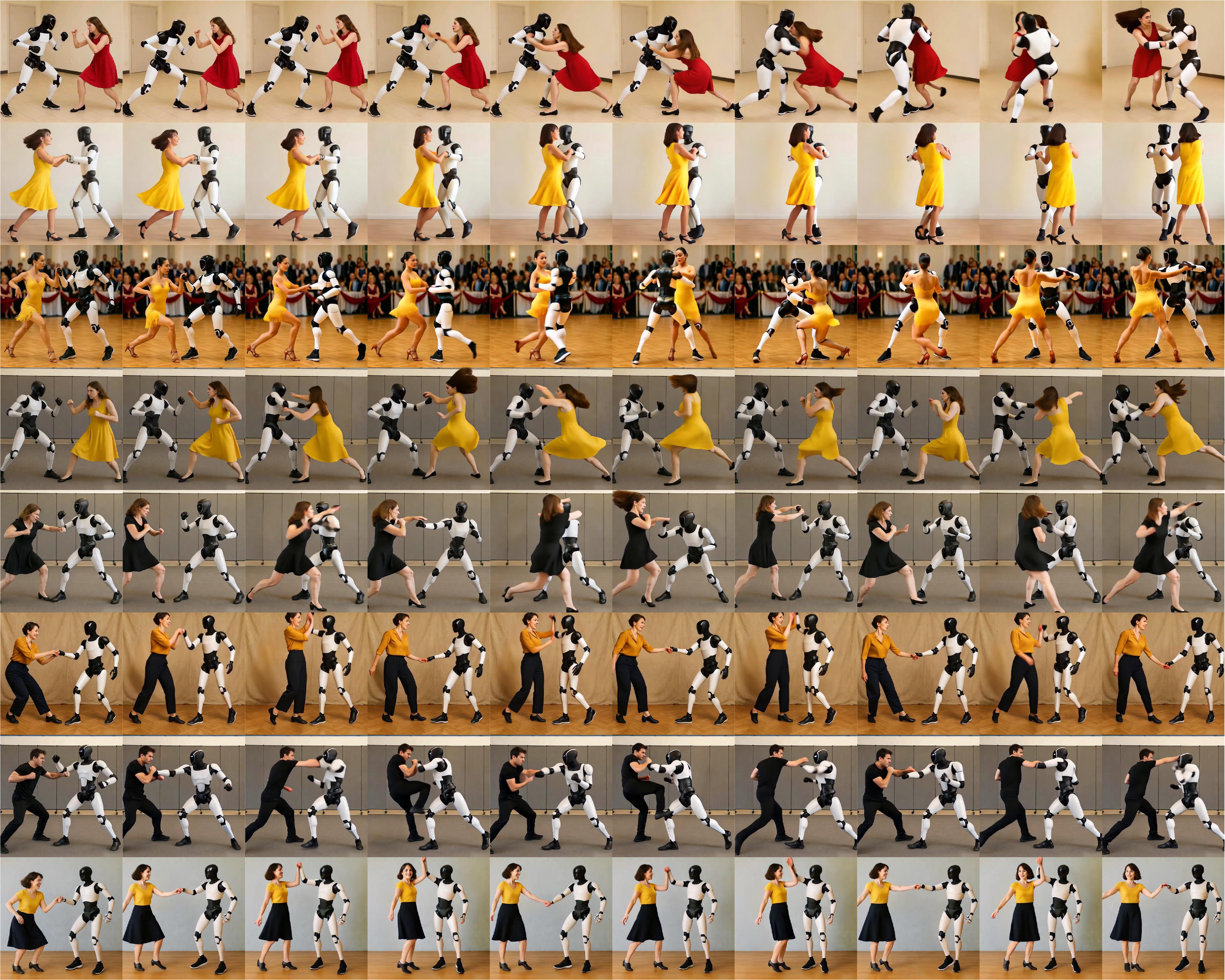}
    \caption{Using the first frame as the reference image, we perform inference on human–robot interaction sequences conditioned on independent pose maps and human masks. }
    \label{fig:rob_results}
\end{figure}

After fine-tuning on our HumanRob-300 humanoid-robot video dataset, \emph{DanceTogether} can generate realistic interaction videos between a humanoid robot and a human (see Fig.~\ref{fig:rob_results}). This demonstrates the effectiveness and generalization ability of \emph{DanceTogether}, offering new insights for embodied-AI and human–robot interaction research.
After the robot and the human exchange positions, both agents retain their original identities. 
The method also handles fine-grained interactions—such as handshakes and sparring—remarkably well.
This part of the video results can be found on the supplementary webpage.

\newpage

\end{document}